\documentclass[a4paper,fleqn]{cas-dc}
\usepackage[numbers,sort&compress]{natbib}
\usepackage[numbers]{natbib}
\usepackage{verbatim} 
\usepackage{amsmath}
\usepackage{comment}
\usepackage{xcolor}
\usepackage{array}
\usepackage{multirow}
\usepackage{amssymb}
\usepackage{amsmath}
\usepackage{multirow}
\usepackage{graphicx}
\usepackage{caption}
\usepackage{cleveref}
\usepackage{booktabs}
\usepackage{times}
\usepackage{caption}
\usepackage{float} 
\usepackage{subcaption}
\usepackage{hyperref}
\usepackage{tabularx}
\usepackage{bbding}
\usepackage{enumitem}
\usepackage[pagewise]{lineno}
\definecolor{ccr}{RGB}{0,0,250}
\hypersetup{hypertex=true,
	colorlinks=true,
	linkcolor=ccr,
	anchorcolor=ccr,
	citecolor=ccr}
	
\begin{document}
\captionsetup[figure]{name={Fig.},labelsep=period} 
\renewcommand{\familydefault}{\rmdefault}
\let\WriteBookmarks\relax
\def\floatpagepagefraction{1}
\def\textpagefraction{.001}
\shortauthors{Y. Ye et~al.}

\title [mode = title]{Radiation, Rotation and Scale Invariant Feature Descriptor for Multimodal Image Matching} 

\author[1]{Yuanxin Ye}[style=chinese,]
\ead{yeyuanxin@home.swjtu.edu.cn}

\author[1]{Tengfeng Tang}[orcid=0000-0002-7709-1882, style=chinese,]
\cormark[1]
\ead{ttf@my.swjtu.edu.cn}
\cortext[cor1]{Corresponding author}

\author[1]{Tao Peng}[style=chinese,]
\ead{pengtao0824@my.swjtu.edu.cn}

\author[1]{Zhiqiang Han}[style=chinese,]
\ead{2021201342@my.swjtu.edu.cn}

\author[2]{Jiayuan Li}[style=chinese,]
\ead{ljy_whu_2012@whu.edu.cn}

\author[3]{Mi Wang}[style=chinese,]
\ead{wangmi@whu.edu.cn}

\affiliation[1]{organization={Faculty of Geosciences and Engineering, Southwest Jiaotong University},
	city={Chengdu}, 
	country={China}}

\affiliation[2]{organization={School of Remote Sensing and Information Engineering, Wuhan University},
	city={Wuhan},
	country={China}}
	
\affiliation[3]{organization={State Key Laboratory of Information Engineering in Surveying, Mapping and Remote Sensing,Wuhan University},
	city={Wuhan},
	country={China}}

\begin{abstract}
Multimodal image matching is a fundamental task for multi-source information fusion. However, geometric distortions and nonlinear radiometric differences (NRD) severely limit matching performance, especially in the presence of radiometric, rotation, and scale variations. To address this issue, we propose a radiation, rotation, and scale invariant (RRSI) feature descriptor for multimodal image matching. First, we design a dual-head regional sampling (DHRS) module, which simultaneously performs Cartesian and Log-Polar sampling on the neighborhoods of keypoints. This parallel encoding strategy enables the RRSI descriptor to retain spatial structural properties while enhancing its robustness to rotation and scale variations. Subsequently, we jointly encode the geometric and radiometric relations between multimodal images within a unified deep feature space, which alleviates the limitation of insufficient receptive fields in the neighborhoods of keypoints, accurately captures key structural features, and achieves feature encoding, interaction, and fusion across intra-modal regions, dual-head sampled regions, and inter-modal regions. Furthermore, we innovatively introduce a cross-modal generative reconstruction constraint during the training phase. By decoding implicit features back into the counterpart modality's structural patches, this bidirectional mechanism explicitly anchors modality-invariant geometric topologies. Ultimately, these components produce an RRSI descriptor that robustly represents common structures under geometric distortions and radiometric differences, thereby improving multimodal image matching accuracy. Experimental results demonstrate that the RRSI achieves highly competitive performance compared to state-of-the-art methods on multimodal image datasets (e.g., optical-infrared and optical-SAR), demonstrating strong robustness to rotation and scale variations. The method supports multimodal matching over the full $[0^\circ,360^\circ]$ rotation range and at scale factors of up to four. Furthermore, the generalization ability of the RRSI has been validated on additional multimodal images, including computer vision, remote sensing, and medical images. The implementation of the proposed method will be made publicly available at https://github.com/yeyuanxin110/RRSI.
\end{abstract}

\maketitle

\section{Introduction}
\label{sec:introduction}
\begin{figure*}
	\includegraphics[width=1\linewidth]{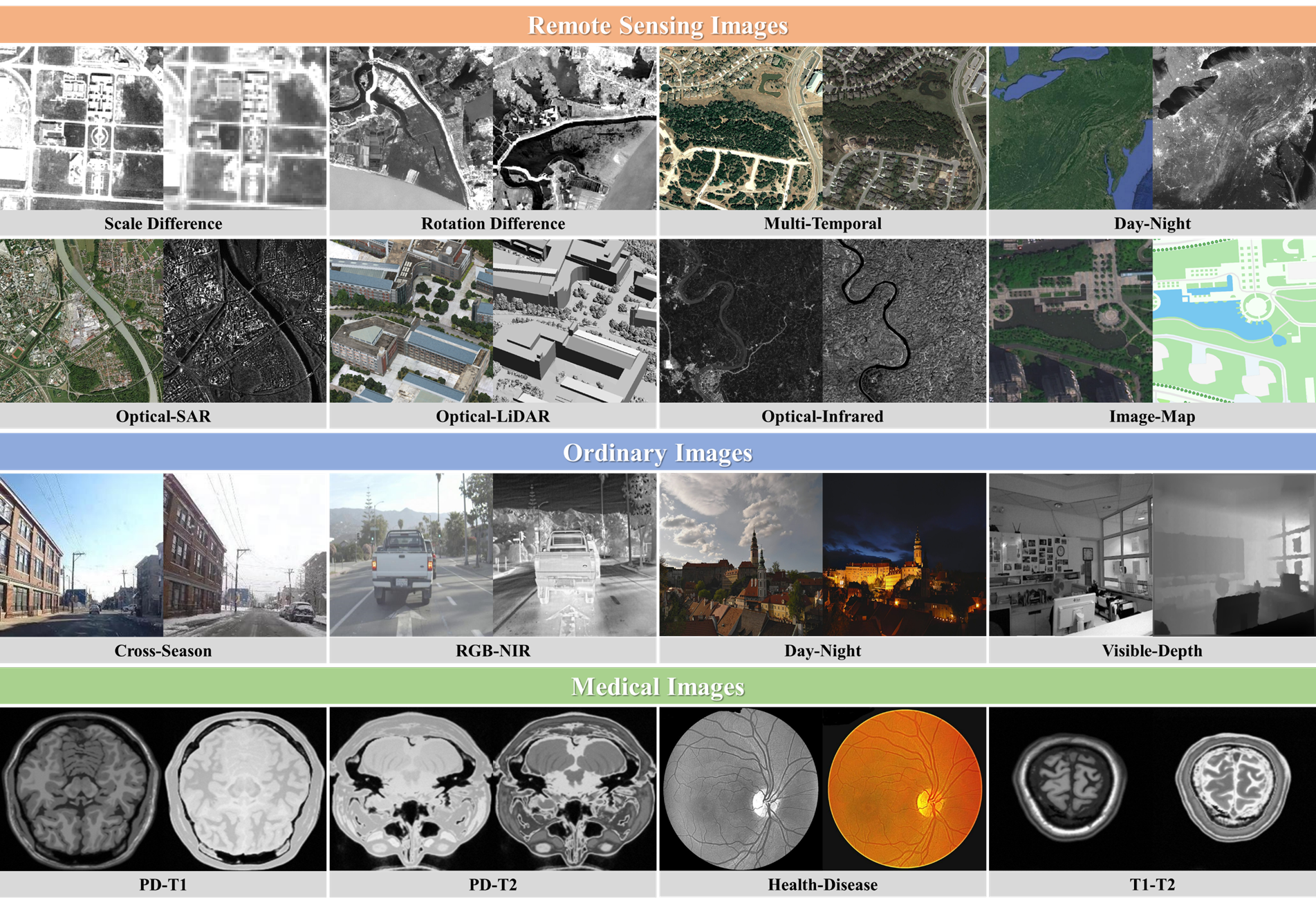}
	\caption{Multimodal image examples from different fields}
	\vspace{-15pt}
	\label{fig:example}
\end{figure*}

Multimodal image matching aims to identify correspondences between two or more images captured by different sensors, at different times, or from different viewpoints \cite{Ma2020ImageMF}, thereby providing spatially aligned data for downstream tasks such as image fusion \cite{karim2023current}, panoramic stitching \cite{fu2023image}, and 3D reconstruction \cite{ji20262d3d}. Multimodal images exhibit not only geometric distortions but also nonlinear radiometric differences (NRD) \cite{zhu2026misr}. Multiple causes contribute to NRD, and one source is variation in external conditions (e.g., multitemporal, day-night, cross-seasonal, and cross-weather imaging). For such data types, the primary sources of matching difficulties include changes in target characteristics, blurring induced by clouds or fog, and variations in ground reflection intensity due to differing lighting conditions. A second source is the difference in physical imaging mechanisms. Within the remote sensing domain, multimodal images encompass optical, infrared, synthetic aperture radar (SAR), light detection and ranging (LiDAR), hyperspectral, and multispectral imaging data, among others. Within the medical domain, multimodal images include computed tomography (CT), magnetic resonance imaging (MRI), and positron emission tomography (PET).

Fig. \ref{fig:example} presents representative examples of the aforementioned multimodal images. Multimodal images deliver information about the same target via different observation methods, facilitating more comprehensive and accurate target interpretation \cite{tang2022superfusion}. For instance, in remote sensing, optical images have rich texture details and high resolution, but their quality is easily affected by weather, making surface information hard to obtain at night or under cloud cover; infrared images rely on thermal radiation to reflect target presence and location, but their target edges are comparatively blurred and their spatial resolution is lower; SAR imaging is unaffected by weather and lighting (able to penetrate clouds and fog) but suffers from significant noise. These multimodal images have distinct strengths, limitations, and complementary characteristics. By integrating their features, the strengths of each data type can be fully leveraged, enhancing the accuracy and efficiency of subsequent information extraction and object interpretation \cite{tang2026robust}. In medical image analysis, anatomical images have high spatial resolution, enabling clear visualization of geometric details (e.g., visceral and skeletal structures) but lack functional information; functional images excel at visualizing metabolic functional changes, yet their clarity is insufficient for revealing structural details, complicating the localization of anatomical structures and boundaries. Thus, combining the complementary information of these two image types is essential to provide more comprehensive patient information and support joint diagnosis and treatment \cite{jiang2021review}. Prior to fusing such complementary information, high-precision multimodal image matching is a prerequisite \cite{tang2022superfusion}.

Existing image matching methods can mainly be divided into two categories: (1) template-based matching methods, and (2) feature-based matching methods. Template matching methods identify correspondences by searching for an extremum of a similarity metric within a template window. These methods usually achieve high accuracy and can be highly efficient in the frequency domain \cite{ye2019fast}. However, if there are significant rotation or scale distortions, prior information is needed to correct these global geometric distortions; otherwise, template matching methods will fail \cite{ye2022multiscale}. Feature matching methods first extract salient features (e.g., points, lines, or regions) using a detection-and-description paradigm, and then determine correspondences by optimizing feature distances. Compared with template matching methods, feature matching methods are more robust to geometric distortion and have broader applicability. The geometric distortion discussed in this paper goes beyond the scope of general template-based matching methods. Therefore, the image matching methods mentioned later mainly refer to feature-based matching methods.

Although image matching techniques have been evolving for decades, multimodal image matching remains challenging. The main difficulties are significant NRD and geometric distortion, which make traditional similarity measures or feature descriptors based on image intensity information almost unusable, leading to a significant decline in matching performance and difficulty in meeting diverse practical requirements. The factors that affect the performance of a matching algorithm mainly include: (1) the repeatability and saliency of keypoints during the feature detection phase; (2) the high consistency of repeatable point features and the discriminability of non-repeatable point features under radiometric, rotational, and scale changes during the feature description phase; (3) the robustness of the matching and outlier rejection model during the feature matching phase. These three aspects constitute major research directions in the field of image matching. There have been many mature studies on the repeatability of keypoint detection \cite{zhu2023r2fd2, li2019rift, zhang2023multilevel,barroso2019key} and the robustness of outlier rejection \cite{chum2005two, wu2014novel,sarlin2020superglue,lindenberger2023lightglue}. However, feature descriptors still pose challenges in terms of radiation, rotation, and scale invariance. The core of multimodal image matching lies in common feature representation, which requires that such descriptors are robust to significant geometric distortions and radiometric differences and can reflect the common properties across modalities. In this paper, the main focus is on common and robust feature description when radiometric, rotational, and scale variations occur simultaneously.

Several urgent issues remain in current multimodal image matching. First, compared with homogeneous image matching, multimodal image matching is more challenging because of significant NRD, making the extraction of common cross-modal features a decisive step. Second, many common features extracted by existing methods remain reliable only for images without significant geometric distortion. Maintaining their consistency under large geometric distortions, particularly when prior transformation information is unavailable, therefore remains an urgent challenge. Solving the image matching problem under large geometric distortion conditions can be further divided into extracting descriptors with rotation and scale invariance. Although feature matching methods are more robust to geometric distortions than template matching methods, existing feature matching methods still struggle with large geometric distortions. Most existing methods solve the multimodal image matching problem under specific conditions, mainly including the following categories: (1) For rotation-invariant feature matching, there are two main solutions: one is to traverse multiple predefined angles and retain the best result through exhaustive matching \cite{li2019rift, zhu2023r2fd2}. The other is to use a small range of rotational perturbations during training to provide the descriptor with limited rotational invariance \cite{detone2018superpoint, tyszkiewicz2020disk, zhao2023aliked}. (2) For scale-invariant feature matching, there are two main solutions: one is to directly extract scale-invariant features from support regions at different scales \cite{ebel2019beyond}, and the other is to increase the set of candidate scale correspondences through image upsampling and downsampling, followed by exhaustive matching in a multiscale space \cite{li2023multimodal}. 

Although current methods demonstrate excellent matching performance in specific scenarios, feature descriptors still face challenges regarding radiation, rotation, and scale invariance. To address these limitations, we innovatively propose a radiation, rotation, and scale-invariant feature descriptor (denoted as RRSI) and construct a multimodal image matching framework based on the RRSI descriptor. The RRSI draws inspiration from the ability of human visual attention to perceive spatial structural similarity across multimodal images. First, to enhance the descriptor’s resistance to geometric distortions (especially large rotation and scale variations), we propose a Dual-Head Regional Sampling (DHRS) module to sample the local neighborhoods around keypoints. Integrating Cartesian and Log-Polar sampling, the DHRS module performs parallel encoding on the acquired dual patches, enabling the RRSI descriptor to retain spatial structural properties while enhancing its robustness to scale and rotation variations. Subsequently, we design a novel RRSI feature description network by jointly encoding geometric and radiometric relations within a unified deep feature space. This encoding process significantly mitigates the limitations imposed by restricted receptive fields in local regions, accurately captures key features and prominent structures in images, and achieves feature encoding, interaction, and fusion across intra-modal regions, dual-head sampling regions, and inter-modal regions. Furthermore, to constrain the feature space against complex modality variations, we innovatively integrate a cross-modal generative reconstruction constraint during the network training phase. By forcing the implicit features to bidirectionally reconstruct the counterpart modality's structural patches, this mechanism explicitly anchors modality-invariant geometric topologies. Ultimately, the RRSI descriptor provides robust invariance to geometric distortions and radiometric differences without imposing an additional computational burden during inference, thereby providing strong support for accurate multimodal image matching.

The main contributions of this paper can be summarized as follows.

\begin{enumerate}[label={\arabic{enumi})}]
	\setlength\itemsep{0em}
	\item We design the DHRS module for local regions around keypoints. Integrating Cartesian and Log-Polar sampling in a dual-branch manner, this module facilitates the RRSI descriptor's learning of spatial structural information while enhancing its robustness to rotation and scale variations.
	\item We propose the RRSI descriptor, which achieves feature encoding, interaction, and fusion across the intra-modal regions, the dual-head sampling regions, and the inter-modal regions by jointly encoding geometric and radiometric relations.
	\item We introduce a training-phase cross-modal generative reconstruction constraint that uses bidirectional structural decoding to regularize the unified feature space and anchor modality-invariant geometric topologies. The auxiliary decoders are removed after training, strengthening descriptor learning without increasing inference-time parameters or computation.
	\item We construct a multimodal image matching framework (the RRSI method) based on the "detection-description-matching" paradigm and conduct extensive experiments on cross-modal datasets (e.g., optical-infrared and optical-SAR). Experimental results demonstrate that the RRSI method achieves superior or highly competitive performance compared with state-of-the-art methods. The RRSI method can adapt to matching tasks involving rotations ranging from $[0^\circ,360^\circ]$ and scale factors of up to four, and its generalization ability has been verified across various cross-modal scenarios in computer vision, remote sensing, and medical imaging.
\end{enumerate}

\section{Related Work}
\label{sec:related_works}
According to the strategy used to establish correspondences, multimodal image matching methods can be generally classified into two categories: (1) template-based matching (also known as area-based matching), and (2) feature-based matching \cite{Ma2020ImageMF,jiang2021review}.
\subsection{Template-Based Matching}
Template-based matching methods first define a template in the sensed image and a search area in the reference image, then adopt a similarity measure (e.g., intensity, probability, or structure) and slide the template across the search area to find an extremum of the similarity measure \cite{ye2017robust}. Structure-based similarity measures (e.g., gradient \cite{heinrich2012mind}, local self-similarity \cite{shechtman2007matching}, and phase consistency \cite{ye2017robust}) are generally more robust to NRD. CFOG \cite{ye2019fast} performs pixel-wise gradient feature representation and utilizes the fast Fourier transform to accelerate template matching. MCGF \cite{zhou2021robust} inputs multiscale and multidirectional gradient features into a CNN, achieving efficient and robust template matching. SPIMNet \cite{ko2023spectral} introduces a domain transformation network and constructs a learnable optical-infrared similarity measure. In general, template matching methods typically exhibit high computational efficiency and matching accuracy, but they are highly sensitive to geometric transformations in images and heavily rely on prior geographic information to correct global geometric distortions, which makes them suitable for image matching scenarios where the transformation relationship between images is nearly translational \cite{ye2022multiscale}.

\subsection{Feature-Based Matching}
\textbf{Handcrafted Features.} Feature-based matching methods aim to extract salient features and establish correspondences between them. They are more robust to geometric distortion \cite{li2019rift}. Beginning with SIFT \cite{lowe2004distinctive}, this paradigm led to major advances in feature matching. However, handcrafted features struggle with multimodal images affected by NRD. To this end, researchers have developed advanced feature descriptors with radiation invariance: RIFT \cite{li2019rift} constructs a radiation-invariant feature descriptor based on phase congruency, WSSF \cite{wan2023multimodal} integrates texture information, Gaussian-Steerable filtering, and edge confidence maps to construct weighted structural saliency features, R2FD2 \cite{zhu2023r2fd2} fuses a repeatable detector and a radiation/rotation invariant descriptor, and SRIF \cite{li2023multimodal} maintains stable matching performance in complex geometric distortion scenarios.

In recent years, deep learning has driven the differentiation of feature matching methods into three major branches: (1) descriptor learning, (2) matcher learning, and (3) multimodal data generation. 

\textbf{Descriptor Learning.} Methods based on descriptor learning focus on optimizing the expressive power of general feature descriptors. For example, HardNet \cite{mishchuk2017working} strengthens feature discriminability through metric learning, SuperPoint \cite{detone2018superpoint} jointly learns keypoint detection and feature description, DISK \cite{tyszkiewicz2020disk} improves local features through reinforcement learning, and ALIKED \cite{zhao2023aliked} captures fine-grained details using deformable kernels. 

\textbf{Matcher Learning.} Methods based on matcher learning focus on building learnable correspondence models: SuperGlue \cite{sarlin2020superglue} optimizes matching probabilities by fusing local and global information through graph neural networks, LightGlue \cite{lindenberger2023lightglue} uses adaptive computation in self-attention and cross-attention layers to reduce computational complexity, LoFTR \cite{sun2021loftr} uses hierarchical Transformers to achieve dense matching without detectors, and its improved version EfficientLoFTR \cite{wang2024efficient} introduces computational optimization strategies. RoMa \cite{edstedt2024roma} refines the features of the DINOv2 \cite{oquab2023dinov2} model and decodes them using Transformers to improve dense matching robustness. XoFTR \cite{tuzcuouglu2024xoftr} introduces a cross-modal feature alignment module based on LoFTR and enhances adaptability to nonlinear radiometric differences through a modality-agnostic feature transformation space. JamMa \cite{lu2025jamma} employs Joint Mamba for efficient local feature matching, while SceneGlue \cite{du2026sceneglue} introduces scene-aware contextual modeling without requiring scene-level annotations.

\textbf{Multimodal Data Generation.} Methods based on multimodal image generation aim to generate large-scale datasets containing multiple modalities, diverse scenarios, and accurate matching labels via a data engine. They then enhance the matching performance of existing general-purpose models in cross-modal scenarios by retraining the models’ weights. For instance, GIM \cite{xuelun2024gim} extracts matching supervision signals from consecutive frames of Internet videos. MatchAnything \cite{he2025matchanything} generates cross-modal images from multiview images and consecutive video frames. MINIMA \cite{ren2025minima} leverages generative models to produce multiple pseudo-modal images from RGB images. Ultimately, these methods greatly expand the quantity and diversity of supervision signals for multimodal images, while significantly improving the cross-modal matching performance of existing models.

Despite these advances, existing methods generally emphasize radiometric robustness, learned correspondence modeling, or data-driven modality expansion in isolation. A unified descriptor that jointly captures cross-modal common structures and remains robust to wide-range rotation and scale variations is still insufficiently explored, particularly when structural regularization is required without retaining auxiliary branches at inference. This gap motivates the RRSI descriptor developed in this work.

\section{Methodology}
\label{sec:method}

\begin{figure*}
	\includegraphics[width=1\linewidth]{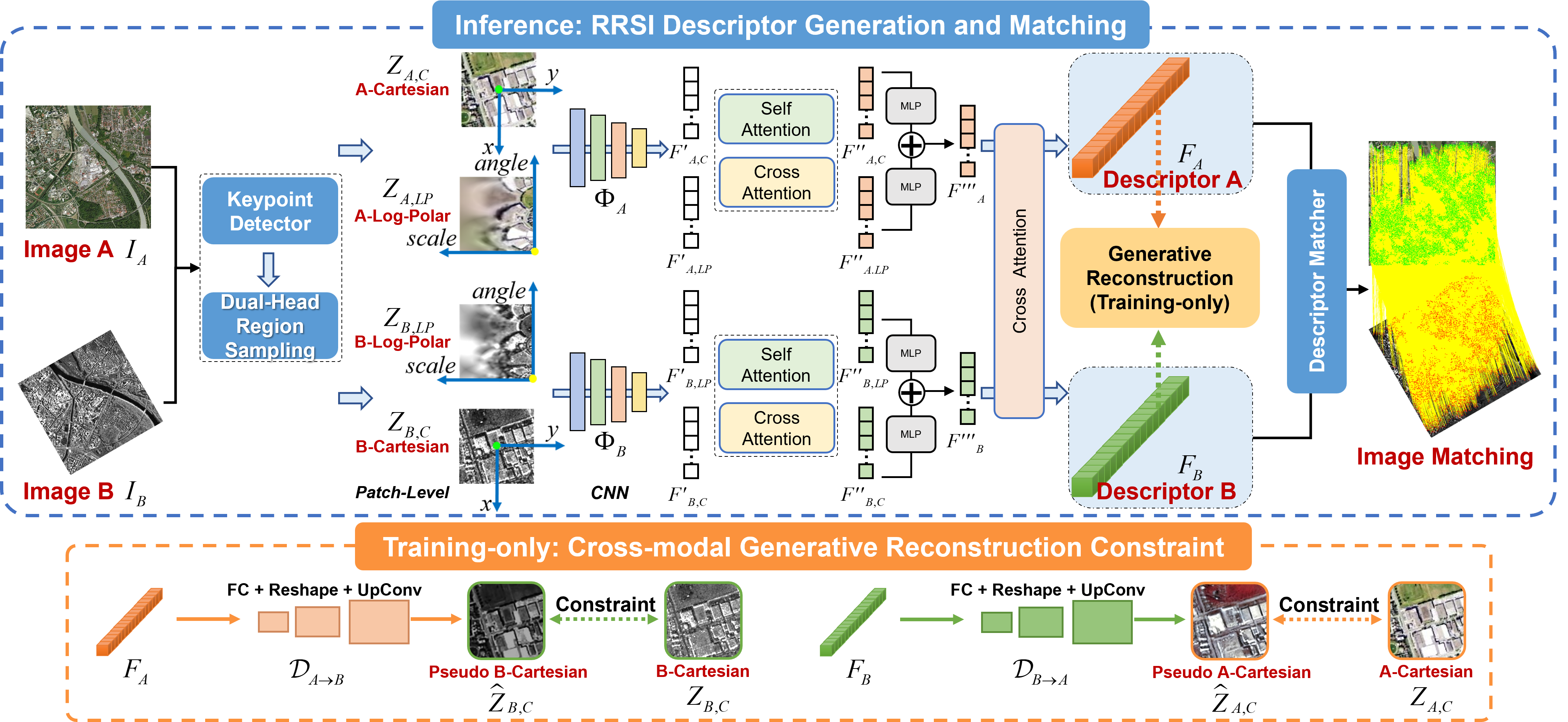}
	\caption{Overall framework of multimodal image matching based on the RRSI descriptor}
	\label{fig:framework}
\end{figure*}

\subsection{Multimodal Image Matching Framework}
\label{sec:framework}
At the core of precise and robust multimodal image matching lies the extraction of common features capable of withstanding geometric distortions and radiometric differences. To this end, we propose the RRSI feature descriptor and construct a matching framework based on it. The matching framework, illustrated in Fig. \ref{fig:framework}, mainly comprises four components: keypoint detection, local region sampling, feature description, and descriptor matching. 

\begin{figure*}[!t]
	\includegraphics[width=1\linewidth]{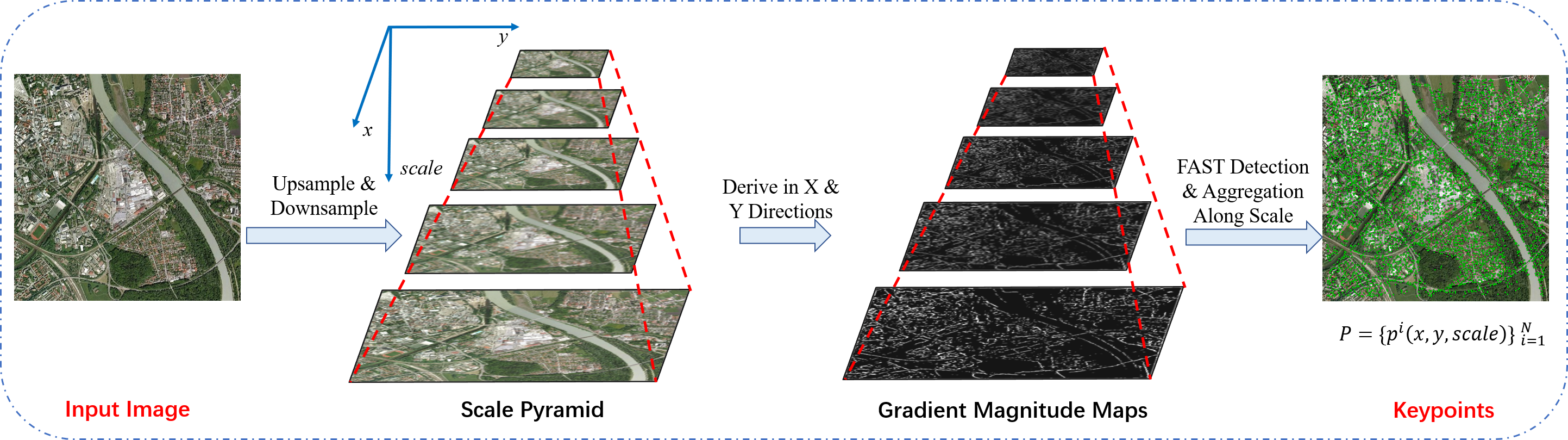}
	\caption{Flowchart of keypoint detection incorporating scale pyramid, gradient magnitude maps, and FAST detector}
	\vspace{-15pt}
	\label{fig:detector}
\end{figure*}

\textbf{Keypoint Detection.} The matching framework is initiated by detecting salient and repeatable keypoints (e.g., corners and edge points) from the input pair of multimodal images. For the input image pair $I_A$ and $I_B$, keypoint detection produces the keypoint sets $P_A=\{p_A^i\}_{i=1}^{N_A}$ and $P_B=\{p_B^i\}_{i=1}^{N_B}$, where $p$ denotes a single keypoint and $N$ denotes the number of keypoints. Specifically, we employ a practical traditional keypoint detection method. As shown in Fig. \ref{fig:detector}, we first construct a scale pyramid for the input image via upsampling and downsampling. The scale pyramid consists of five layers, with scaling ratios of $[1/2, \sqrt{1/2}, 1, \sqrt{2}, 2]$ relative to the original image, respectively. This step improves the robustness of subsequent matching to scale variations, and the scale pyramid also improves robustness to image degradation (e.g., noise, occlusion, and deformation). Then, we compute partial derivatives in the $x$ and $y$ directions for each scale level, take their absolute values, and thereby obtain the corresponding gradient magnitude maps:
\begin{equation}
	G_s=\left| {\partial I_s}/{\partial x} \right| + \left| {\partial I_s}/{\partial y} \right|
\end{equation} 
where \(I_s\) and \(G_s\) denote the image and gradient magnitude map at the $s$-th scale level, respectively. Compared with the original image, the gradient magnitude map can more effectively represent edge and corner information. In particular, when cross-modal images exhibit significant radiometric differences, keypoint detection using gradient magnitude provides greater saliency and repeatability. Finally, we employ the FAST detector to identify keypoints on gradient magnitude maps across the scale pyramid and aggregate these keypoints along the scale dimension, resulting in the keypoint set $P=\{p^i(x,y,\mathrm{scale})\}^N_{i=1}$.

\textbf{Local Region Sampling.} Local region sampling extracts image neighborhoods centered on the keypoints. To enhance the robustness of image matching to geometric distortions, particularly large rotations or scale variations, we innovatively propose the DHRS module, which incorporates both Cartesian sampling \(T_c\) and Log-Polar sampling \(T_{lp}\). This process can be expressed as:
\begin{equation}
	\label{equ:DHRS}
	z_C^i=T_c(p^i), \qquad z_{LP}^i=T_{lp}(p^i)
\end{equation} 
where \(z_C^i\) and \(z_{LP}^i\) denote the independently sampled Cartesian and Log-Polar patches corresponding to keypoint \(p^i\), respectively. For image \(I_A\), the DHRS module produces two independent patch sets, \(Z_{A,C}=\{z_{A,C}^i\}_{i=1}^{N_A}\) and \(Z_{A,LP}=\{z_{A,LP}^i\}_{i=1}^{N_A}\), both in \(\mathbb{R}^{N_A\times128\times128}\). They are compactly denoted as the ordered pair \(Z_A=(Z_{A,C},Z_{A,LP})\). The corresponding sets \(Z_{B,C}\) and \(Z_{B,LP}\) are generated for image \(I_B\) in the same manner. The details of the DHRS module will be elaborated in Section \ref{sec:DHRS}.

\textbf{Feature Description.} Feature description abstracts image information from local regions into one-dimensional feature vectors that describe the corresponding keypoints. We innovatively propose the RRSI descriptor. As shown in Fig. \ref{fig:framework}, the RRSI description network jointly models geometric and radiometric relations within a unified deep feature space. It is not limited to extracting convolutional features from individual local regions, but further achieves deep fusion and interaction among intra-modal regions, the dual-head sampling branches, and cross-modal image regions. Following feature extraction, we obtain two descriptor sets \(F_A = \{f^i_A\}^{N_A}_{i=1}\) and \(F_B = \{f^i_B\}^{N_B}_{i=1}\), where $f$ denotes a single descriptor. Each keypoint \(p^i\) corresponds to an independently sampled patch pair \((z_C^i,z_{LP}^i)\) and one feature descriptor \(f^i\). The inference and training processes of the RRSI feature descriptor are described in Section \ref{sec:RRSI} and Section \ref{sec:loss}, respectively.

\textbf{Descriptor Matching.} After obtaining the RRSI descriptors, we utilize the normalized Euclidean distance between descriptors and efficiently search for potential correspondences using the FLANN \cite{muja2009fast} algorithm. During this process, multiscale keypoints and their descriptors from the scale pyramid are mapped back to the original image before descriptor matching. Ultimately, this enables accurate and robust matching between multimodal images \(I_A\) and \(I_B\).

\subsection{Dual-Head Region Sampling}
\label{sec:DHRS}
Following keypoint detection, a critical step involves selecting an appropriate neighborhood around each keypoint and sampling it into patches for subsequent feature description and matching. One common approach is Cartesian sampling \(T_c\), in which the neighboring region of a keypoint is cropped and resampled \cite{mishchuk2017working,tian2020hynet,xu2023sar}. This sampling mode fully preserves the spatial structure around the keypoint, and \(T_c\) in \eqref{equ:DHRS} can be expressed as:

\begin{equation}
	\begin{bmatrix}  x^{p} \\  y^{p}\end{bmatrix}=\begin{bmatrix} ({W^{p}}/{W^{i}})\cos \theta  & -({H^{p}}/{H^{i}})\sin \theta & x_{0}^{i}\\ ({W^{p}}/{W^{i}})\sin \theta  & ({H^{p}}/{H^{i}})\cos \theta  & y_{0}^{i}\end{bmatrix}\begin{bmatrix}  x^{i} \\  y^{i} \\ 1\end{bmatrix}
\end{equation} 
where $(x_0^i,y_0^i)$ represents a keypoint, $(x^i,y^i)\to (x^p,y^p)$ represents the coordinate transformation of a neighboring point from the image to a new patch, and $\theta$ represents the orientation of a keypoint. If a keypoint does not have an assigned orientation, then $\theta$ is set to 0.

\begin{figure}
	\centering
	\includegraphics[width=1\columnwidth]{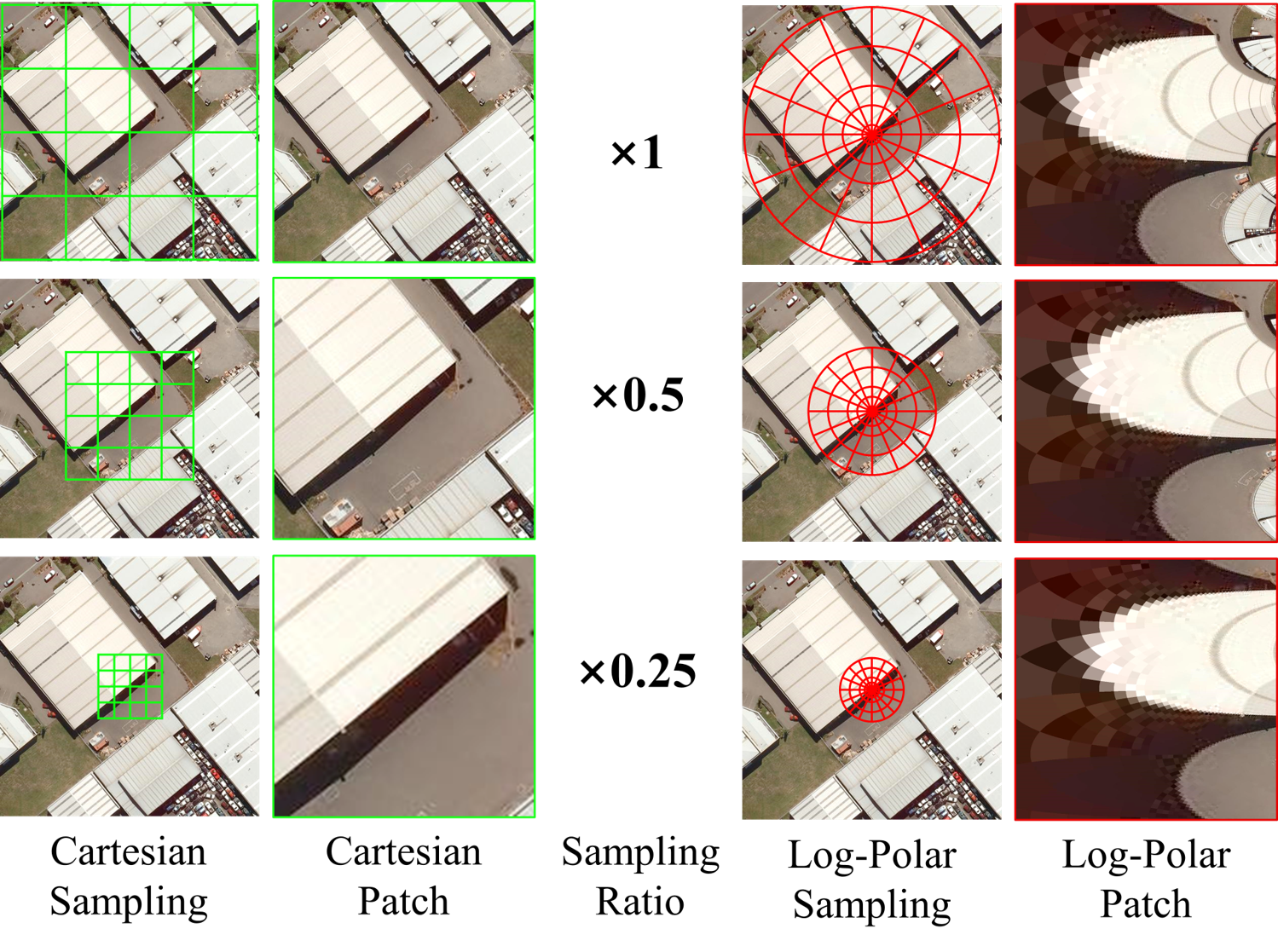} 
	\caption{Local region sampling (Cartesian vs. Log-Polar) under different sampling ratios}
	\label{fig:sampling} 
	\vspace{-15pt}
\end{figure}

However, the content of Cartesian patches undergoes significant variations under different sampling ratios. The sampling ratio is defined as the ratio of patch range $W^{p}/ H^{p}$ to image range $W^{i} / H^{i}$. As illustrated in the left part of Fig. \ref{fig:sampling}, Cartesian sampling employs a uniform sampling manner, where redundant information at the edges of a patch shares the same density as the central region. When the sampling ratio is scaled to 0.5 times, the overlapping proportion between two Cartesian patches only account for 25\%. As the scale difference increases, the overlapping proportion will further decrease.

If only these Cartesian patches are fed into the subsequent learning network, expecting the descriptors to achieve scale invariance, the model would struggle to ensure accurate feature matching \cite{ebel2019beyond}. To address the impact of scale distortion on patch content similarity and descriptor correlation, a more uneven strategy, Log-Polar sampling \(T_{lp}\) can be adopted. \(T_{lp}\) in \eqref{equ:DHRS} can be further expressed as: 
\begin{equation}
	\begin{aligned}
		&x^i = x^i_0 + e^{\log (r)  x^p/W^i} \cos (2\pi y^p/H^i),\\
		&y^i = y^i_0 + e^{\log (r)  x^p/W^i} \sin (2\pi y^p/H^i)
	\end{aligned}
\end{equation} 
where $r$ is the max sample radius centered on the keypoint $(x^i_0,y^i_0)$.

Log-Polar sampling is illustrated on the right side of Fig. \ref{fig:sampling}. Compared with Cartesian sampling, it has three distinctive properties: (1) Rotation equivariance: tangential rotation changes around the center under Cartesian coordinates correspond to displacement on the vertical axis under Log-Polar coordinates; (2) Scale equivariance: radial scale changes around the center under Cartesian coordinates correspond to displacement on the horizontal axis under Log-Polar coordinates; (3) Uneven sampling: points near the sampling center have a higher sampling density, while less informative areas far from the center are sampled more sparsely, which facilitates the extraction of scale-invariant features.

The proposed DHRS module performs both Cartesian and Log-Polar sampling simultaneously. The dual patches obtained by the DHRS module are then encoded in parallel by the subsequent RRSI description network, thereby enhancing the resistance of the RRSI descriptor to scale and rotation variations. The ablation experiment on the DHRS module will be conducted in Section \ref{sec:ablation study}.

\subsection{Radiometric, Rotational, Scale-Invariant Feature Descriptor}
\label{sec:RRSI}
The core of multimodal image matching lies in common feature representation, which requires that such descriptors are robust to significant geometric distortions and radiometric differences and can reflect the common properties across modalities. Therefore, we innovatively propose the RRSI descriptor for common feature representation. 

Typically, convolutional neural networks (CNNs) are used to gradually abstract image patches into deep features. However, extracting convolutional features solely from local regions restricts the feature receptive field and impedes sufficient information interaction, which undermines the performance of subsequent feature matching. The advent of attention mechanisms addresses this issue by facilitating feature interaction and modeling dependencies among local-region features during extraction. Specifically, the network not only focuses on areas adjacent to local regions but also enables interaction between local regions and other distant regions within the same image, or even cross-image regions. This achieves adequate feature fusion and significantly mitigates the limitation of a restricted receptive field. Inspired by the human visual attention system that naturally identifies structural similarities across multimodal images, and building on advances in ViT \cite{dosovitskiy2020image} and Cross-ViT \cite{chen2021crossvit}, we design the RRSI description network by jointly encoding the geometric and radiometric relations within a unified deep feature space.

\begin{figure*}[!t]
	\includegraphics[width=1\linewidth]{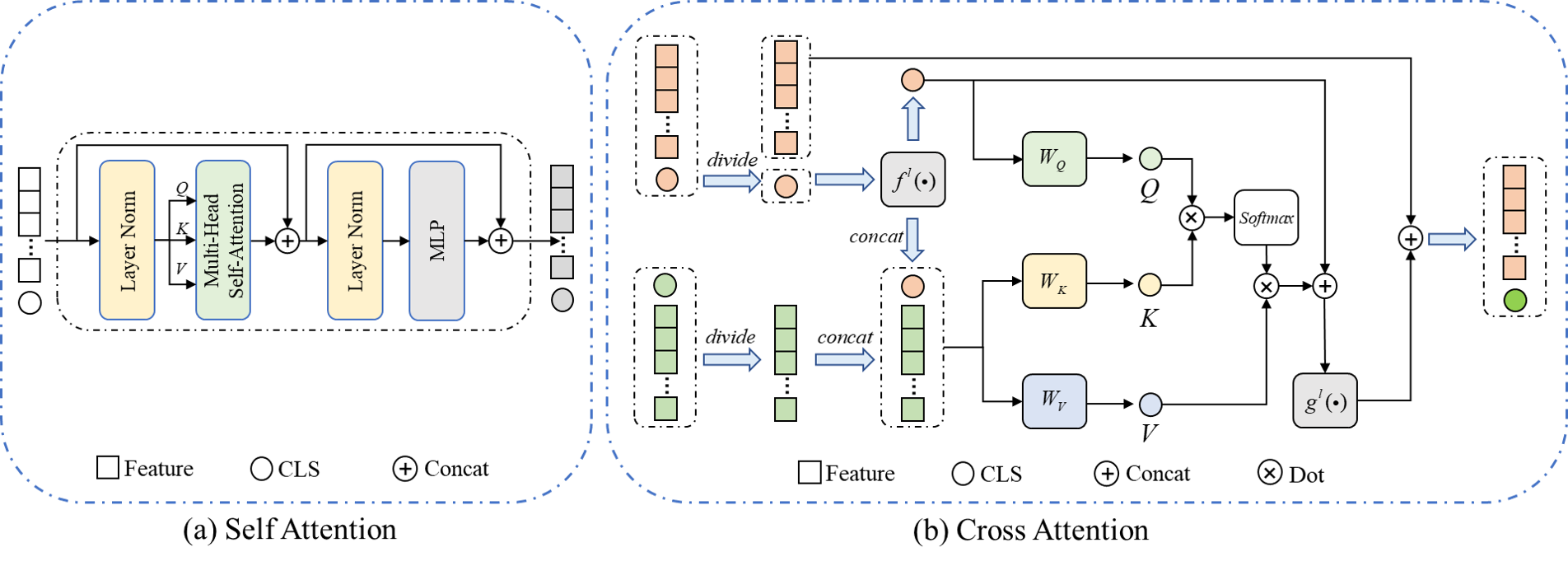}
	\caption{The architecture of self attention and cross attention}
	\vspace{-15pt}
	\label{fig:attention}
\end{figure*}

Specifically, taking image \(I_A\) with modality $A$ as an example: keypoint detection is performed to obtain the keypoint set \(P_A \in \mathbb{R}^{N_A \times 2}\), and the ordered pair of independent patch sets \(Z_A=(Z_{A,C},Z_{A,LP})\) is directly generated by the DHRS module, where \(Z_{A,C},Z_{A,LP}\in\mathbb{R}^{N_A\times128\times128}\). Subsequently, a pseudo-siamese VGG network is employed to extract deep convolutional features from \(Z_{A,C}\) and \(Z_{A,LP}\), respectively, yielding the initial feature sets \(F'_{A,C} \in \mathbb{R}^{N_A \times 256}\) and \(F'_{A,LP} \in \mathbb{R}^{N_A \times 256}\).

Next, within the intra-modal setting, \(F'_{A,C}\) and \(F'_{A,LP}\) undergo \(k_1\) self-attention layers in their respective feature spaces. Subsequently, in a unified feature space, they pass through \(k_2\) cross-attention layers, resulting in the intra-modal refined feature sets \(F''_{A,C} \in \mathbb{R}^{N_A \times 256}\) and \(F''_{A,LP} \in \mathbb{R}^{N_A \times 256}\). Following this, \(F''_{A,C}\) and \(F''_{A,LP}\) each pass through a single MLP layer and are then summed to yield the fused feature \(F'''_{A} \in \mathbb{R}^{N_A \times 256}\). As shown in Fig. \ref{fig:attention}, self-attention enables feature interaction within a single branch, achieving information transfer across local regions via the combination of Layer Normalization, multi-head attention, and MLP. Cross-attention, on the other hand, facilitates feature fusion between two distinct branches in a unified space, completing inter-branch information integration through the interaction mechanism. The objectives of this step are twofold: First, since the initial features \(F'_{A,C}\) and \(F'_{A,LP}\) are derived from local-region convolutions with limited receptive fields, the self-attention mechanism enables feature interaction across different local regions. This expands the feature scope beyond individual local regions to incorporate other regions, thereby enlarging the receptive field. Second, since \(F'_{A,C}\) (Cartesian-sampled) emphasizes patch spatial structure and \(F'_{A,LP}\) (Log-Polar-sampled) focuses more on rotation/scale resistance, cross-attention-based interaction and fusion between their features facilitates learning enhanced representations, retaining spatial information while boosting geometric distortion robustness.

While the above steps describe the intra-modal feature encoding, interaction, and fusion process using Image \(I_A\) with modality $A$ as an example, the same procedure is applied to Image \(I_B\) with modality $B$ but without sharing weights with the network for modality $A$, resulting in the fused feature \(F'''_B \in \mathbb{R}^{N_B \times 256}\). Next, within the inter-modal setting, the fused feature sets \(F'''_A\) (from image A) and \(F'''_B\) (from image B) pass through \(k_3\) cross-attention layers in a unified feature space. Since \(F'''_A\) and \(F'''_B\) are derived from cross-modality image features and significant non-linear radiometric differences exist between cross-modality images, this step aims to facilitate further alignment of cross-modality features in the deep space via interacting features from different modalities. It prioritizes shared feature information across cross-modality images, reduces redundant information interference on subsequent feature matching, and enhances feature robustness to radiometric differences. Ultimately, after the final cross-attention layer, the feature descriptor sets $F_A$ and $F_B$ are obtained respectively, which are the RRSI descriptors.

\begin{figure*}[!t]
	\centering
	\includegraphics[width=0.95\linewidth]{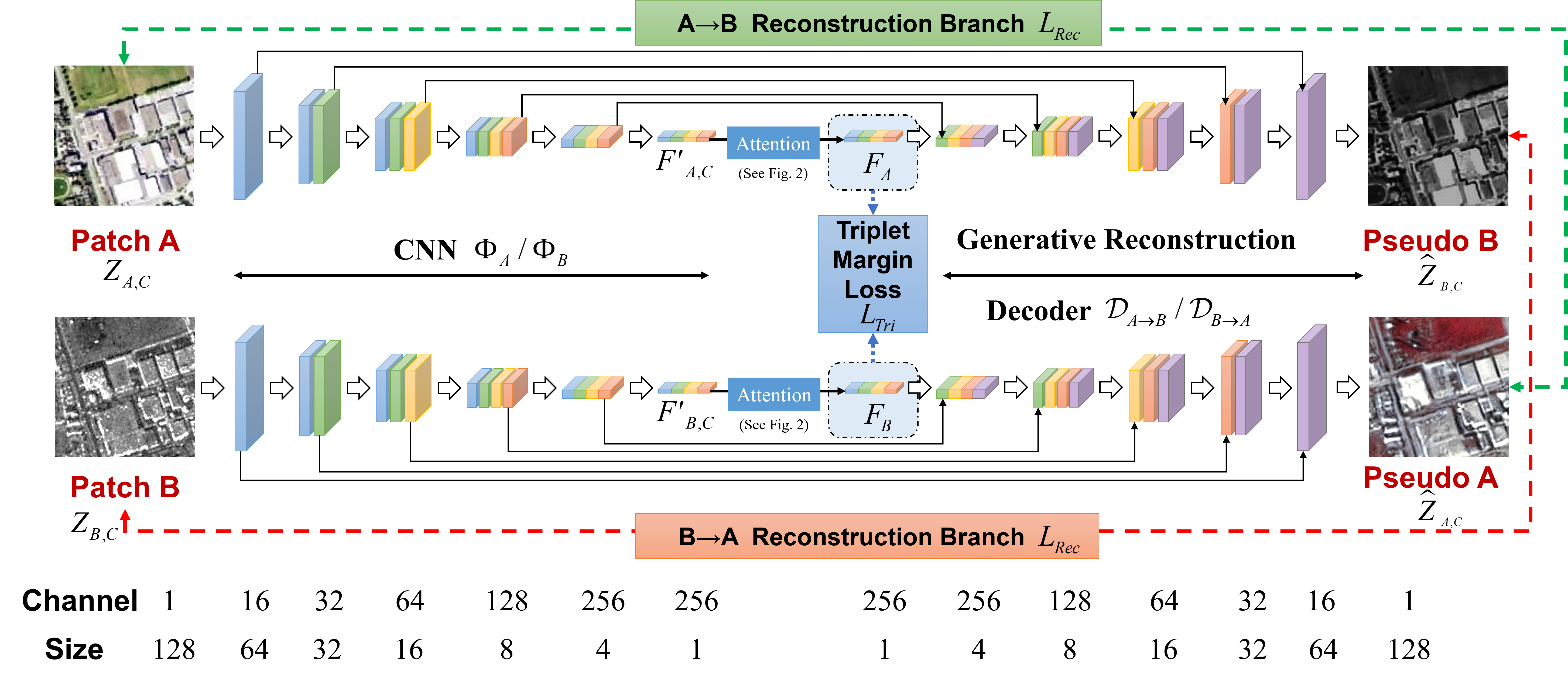}
	\caption{Training-phase supervision mechanism of the cross-modal generative reconstruction decoders. The two symmetric branches illustrate cross-modal pseudo-patch generation and the source-side skip connections used to regularize the joint representation space through $L_{Rec}$.}
	\label{fig:decoder_details}
\end{figure*}

\subsection{Training-Phase Generative Reconstruction and Loss Functions}
\label{sec:loss}

Multimodal images exhibit significant geometric distortions (e.g., rotation, scale variations) and non-linear radiometric differences, resulting in significant deviations in the feature distribution of cross-modal matching regions. Relying solely on raw patch pairs makes it difficult for the model to learn robust common feature representations. Therefore, during the training phase, we construct a comprehensive multi-objective optimization strategy.

\subsubsection{Triplet Margin and Regularization Constraints}
We select the triplet margin loss \cite{mishchuk2017working} as the foundational constraint for training the RRSI descriptor. The triplet training strategy minimizes the distances between matched descriptors and maximizes the distances between unmatched descriptors.

During the training process, taking images \(I_A\) and \(I_B\) as examples, $N$ distinct points \(P_A = \{p_A^i\}_{i=1}^N\) are randomly extracted from \(I_A\). Meanwhile, based on the supervision label (e.g., the geometric transformation relationship \(H_{AB}\) between images \(I_A\) and \(I_B\)), the corresponding points \(P_B = \{p_B^i\}_{i=1}^N\) are obtained. Using the DHRS module, the ordered pairs of independent patch sets \(Z_A=(Z_{A,C},Z_{A,LP})\) and \(Z_B=(Z_{B,C},Z_{B,LP})\) are obtained. Further, through the RRSI feature description network, the corresponding descriptor sets \(F_A = \{f_A^i\}_{i=1}^N\) and \(F_B = \{f_B^i\}_{i=1}^N\) are obtained. Under the triplet training strategy, we regard the $N$ pairs of matching region features as positive sample pairs $P^+$, while all non-matching region features are considered as \(N(N-1)\) negative sample pairs $P^-$, which can be expressed as:

\begin{equation}
	\label{equ:sample}
	\begin{aligned}
		&P^+ = \{(f_A^i, f_B^i)\mid  i=1,2,...,N \},\\
		&P^- = \{(f_A^i, f_B^j) \mid i=1,2,...,N, j=1,2,...,N, i \neq j\}
	\end{aligned}
\end{equation} 

We adopt a hybrid distance \cite{tian2020hynet} instead of the Euclidean distance between two feature descriptors $f_A^i$ and $f_B^j$ to form the objective function of the RRSI. The hybrid distance enhances the consistency of matching sample pairs and reduces information loss parallel to the feature direction, which can be expressed as:

\begin{equation}
	\label{equ:distance}
	D_h(f_A^i, f_B^j) = 2 - 2\frac{f_A^i \cdot f_B^j}{\|f_A^i\| \, \|f_B^j\|} + \sqrt{2\left(1 - \frac{f_A^i \cdot f_B^j}{\|f_A^i\| \, \|f_B^j\|}\right)}
\end{equation}

Given the hybrid distance, \( N \) hardest negative sample pairs \( P^-_{\text{hardest}} \) are selected from \( P^- \). \( P^-_{\text{hardest}} \) contains the most challenging negative samples for model training, which can effectively drive the model to learn more discriminative feature representations. Specifically, for each anchor feature \( f_A^i \) from image \( I_A \), the hardest negative sample is the feature \( f_B^j \) from image \( I_B \) that minimizes the hybrid distance \( D_h(f_A^i, f_B^j) \) among all non-matching pairs. This is formulated as:
\begin{equation}
	\label{equ:hardest}
	\begin{aligned}
		&j^*(i) = \underset{j \neq i}{\arg\min} \, D_h(f_A^i, f_B^j), \\
		&P^-_{\text{hardest}} = \left\{ \left(f_A^i, f_B^{j^*(i)}\right) \mid i=1,2,\ldots,N \right\}
	\end{aligned}
\end{equation}

Based on the positive sample pairs $P^+$ in \eqref{equ:sample}, the hardest negative sample pairs in \eqref{equ:hardest}, and the hybrid distance in \eqref{equ:distance}, the triplet margin loss is calculated as follows:

\begin{equation}
	\label{equ:triplet}
	L_{Tri} = \frac{1}{N} \sum_{i=1}^{N} \max\left(0, \, m + D_h(P^+(i)) - D_h(P^-_{hardest}(i))\right)
\end{equation}
where $m$ is the triplet margin that controls the distance boundary between positive and negative samples.

We also add a similarity regularization term to constrain the descriptors and improve model robustness to image-intensity and radiometric changes, which can be expressed as:
\begin{equation}
	\label{equ:reg}
	L_{Reg} = \frac{1}{N} \sum_{i=1}^{N} (|f_A^i|-|f_B^i|)^2
\end{equation}

\subsubsection{Cross-Modal Generative Reconstruction Constraint}
To further regularize the unified deep feature space and prevent potential feature drift under extreme NRD, we innovatively introduce a cross-modal generative reconstruction mechanism that operates exclusively during the training phase. Specifically, two lightweight structural decoders, denoted as $\mathcal{D}_{A\to B}$ and $\mathcal{D}_{B\to A}$, are appended to the trailing edge of the inter-modal cross-attention layers.

As illustrated in Fig. \ref{fig:decoder_details}, the highly abstracted implicit descriptors $F_A$ and $F_B$ serve as the principal semantic inputs to the decoders. To recover spatial details progressively, the decoders also receive source-side multi-scale CNN features, denoted as $\mathcal{S}_A=\{S_A^{(l)}\}_{l=1}^{L}$ and $\mathcal{S}_B=\{S_B^{(l)}\}_{l=1}^{L}$ for modalities $A$ and $B$, respectively. Symmetrically, the bidirectional generation process can be formulated as:
\begin{equation}
	\begin{aligned}
	\hat{Z}_{B,C} &= \mathcal{D}_{A\to B}(F_A,\mathcal{S}_A),\\
	\hat{Z}_{A,C} &= \mathcal{D}_{B\to A}(F_B,\mathcal{S}_B)
	\end{aligned}
\end{equation}
Thus, the $A\to B$ decoder receives only the descriptor and skip features originating from modality $A$, whereas the $B\to A$ decoder receives only those originating from modality $B$; neither decoder accesses features from the target-modality encoder. Structurally, each decoder is implemented using a series of transposed convolutional layers paired with instance normalization and ReLU activations, capped with a Tanh function. 

This generative mapping establishes a dual-level regularization. First, to enforce spatial and geometric layout fidelity during cross-modal reconstruction, a pixel-level reconstruction loss $L_{pix}$ is defined:
\begin{equation}
	\label{equ:loss_pix}
	L_{pix} = \frac{1}{N} \sum_{i=1}^{N} \left( \| Z_{B,C}^i - \hat{Z}_{B,C}^i \|_1 + \| Z_{A,C}^i - \hat{Z}_{A,C}^i \|_1 \right)
\end{equation}

Second, to prevent feature distribution drift and ensure that the reconstructed patches retain consistent high-level semantics, a feature-level perceptual consistency loss $L_{feat}$ penalizes semantic deviations by constraining the re-projected features against the original Cartesian CNN features ($F_{A,C}^{\prime i}$ and $F_{B,C}^{\prime i}$) prior to the attention layers. Here, $\Phi_A$ and $\Phi_B$ denote the modality-specific Cartesian CNN encoders, such that $F'_{A,C}=\Phi_A(Z_{A,C})$ and $F'_{B,C}=\Phi_B(Z_{B,C})$. Each operator shares weights with the corresponding encoder used in forward feature extraction; however, these encoder parameters are frozen when the reconstruction loss is computed. The original features are further treated as fixed supervision targets through the stop-gradient operator $\operatorname{sg}(\cdot)$. Consequently, $L_{feat}$ does not update the target features or encoder parameters, while gradients through $\Phi_A(\hat{Z}_{A,C})$ and $\Phi_B(\hat{Z}_{B,C})$ with respect to the reconstructed patches remain available to optimize the decoders and the trainable descriptor pathway:
\begin{equation}
	\label{equ:loss_feat}
	\begin{aligned}
	L_{feat} = \frac{1}{N} \sum_{i=1}^{N} \Big(&\| \operatorname{sg}(F_{A,C}^{\prime i}) - \Phi_A(\hat{Z}_{A,C}^i) \|_2^2 \\
	&+ \| \operatorname{sg}(F_{B,C}^{\prime i}) - \Phi_B(\hat{Z}_{B,C}^i) \|_2^2 \Big)
	\end{aligned}
\end{equation}

The total auxiliary generative reconstruction loss $L_{Rec}$ is formulated as:
\begin{equation}
	\label{equ:loss_rec}
	L_{Rec} = L_{pix} + \alpha L_{feat}
\end{equation}
where $\alpha$ is a hyperparameter balancing the two reconstruction tiers. Ultimately, the comprehensive multi-objective loss function $L_{total}$ is summarized as a linear combination of the structural contrastive, regularization, and generative terms:
\begin{equation}
	\label{equ:loss_total}
	L_{total} = L_{Tri} + \beta L_{Reg} + \gamma L_{Rec}
\end{equation}
where $\beta$ and $\gamma$ are trade-off hyperparameters. 

Because corresponding multimodal patches may differ substantially in radiometric appearance, $L_{pix}$ is neither expected to converge to zero nor intended to achieve exact cross-modal image translation. Instead, it serves as an auxiliary structural constraint, and its contribution through $L_{Rec}$ is controlled by $\gamma$. Once joint training is complete, both decoder branches are removed, so this constraint introduces no additional parameters or computation during image matching inference.

\subsection{Technical Details}
\label{sec:details}

\textbf{Network Settings.} During the CNN encoding of local regions, we employ the VGG-16 network and adapt it to a 128×128 input dimension and a 1×256 output feature vector. During the intra-modal attention, the number of self-attention layers \(k_1\) is set to 2, and the number of cross-attention layers \(k_2\) is also set to 2. During the inter-modal attention, the number of cross-attention layers \(k_3\) is set to 2. Each attention module is configured with four heads and maintains an input and output dimension of 1×256. Furthermore, the newly introduced cross-modal generative decoders $\mathcal{D}_{A\to B}$ and $\mathcal{D}_{B\to A}$ are symmetrically constructed using a cascade of transposed convolutional layers with source-side multi-scale skip connections from the CNN encoders to progressively upsample the highly compressed semantic descriptors back to the dense spatial resolution.
The symmetric decoding process, progressive feature-scale transitions, and source-side skip connections are illustrated in Fig. \ref{fig:decoder_details}.

\textbf{Matching Framework Settings.} The upper limit on the number of detected keypoints per image is set to 5000. During the inference phase, for the DHRS module, the local region sampling size is fixed at 128×128 pixels. 

\textbf{Training Settings.} The training hyperparameters are established as follows: the margin $m$ for the triplet loss is set to 1.2; for the multi-objective loss function, the trade-off weights are empirically set to $\beta = 0.1$ (for similarity regularization), $\gamma = 0.5$ (for the total generative reconstruction), and $\alpha = 0.1$ (to balance the feature-level perceptual consistency within the reconstruction). The training batch size $N$ is 300. Data augmentation is performed using Kornia \cite{riba2020kornia}, including intensity transformations (e.g., brightness, contrast, blur, and noise) and geometric transformations (e.g., rotation, shear, and flip). Random rotation angles span the full range of $[0^\circ,360^\circ]$, and the scaling augmentation range is [0.75, 1.5]. Specifically, when the DHRS module samples local regions, using 128×128 pixels as the reference size, patches ranging from 96×96 to 192×192 pixels are randomly sampled from the original training images. These patches are then dynamically resampled to 128×128 pixels to satisfy the network's input dimensional constraints. During the training phase, random points are utilized in lieu of detected keypoints, with the precise correspondences of these random points determined via the ground-truth geometric transformation matrix \(H_{AB}\) of the multimodal image pairs.

\section{Experiments} 
\label{sec:experiments}
To validate the matching performance of the proposed RRSI method, the experimental setting is first elaborated in Section \ref{sec:experimental setting}, including introductions to datasets, comparative methods, evaluation metrics, and implementation details. Next, in Section \ref{sec:image matching}, comparative experiments are conducted between the RRSI method and nine representative methods. Subsequently, in Section \ref{sec:rotate invariance}, the applicability of the RRSI method to rotation-invariant matching is assessed. In Section \ref{sec:scale invariance}, its applicability to scale-invariant matching is evaluated. Furthermore, Section \ref{sec:ablation study} verifies the effectiveness of the core modules within the RRSI framework. Finally, Section \ref{sec:generalization} examines the transferability of the RRSI method to unseen multimodal image types.

\subsection{Experimental Setting}
\label{sec:experimental setting}

\begin{figure}
	\centering
	\subfloat[LLVIP (Visible-Infrared) \label{fig:dataset_llvip}]{\includegraphics[width=1\linewidth]{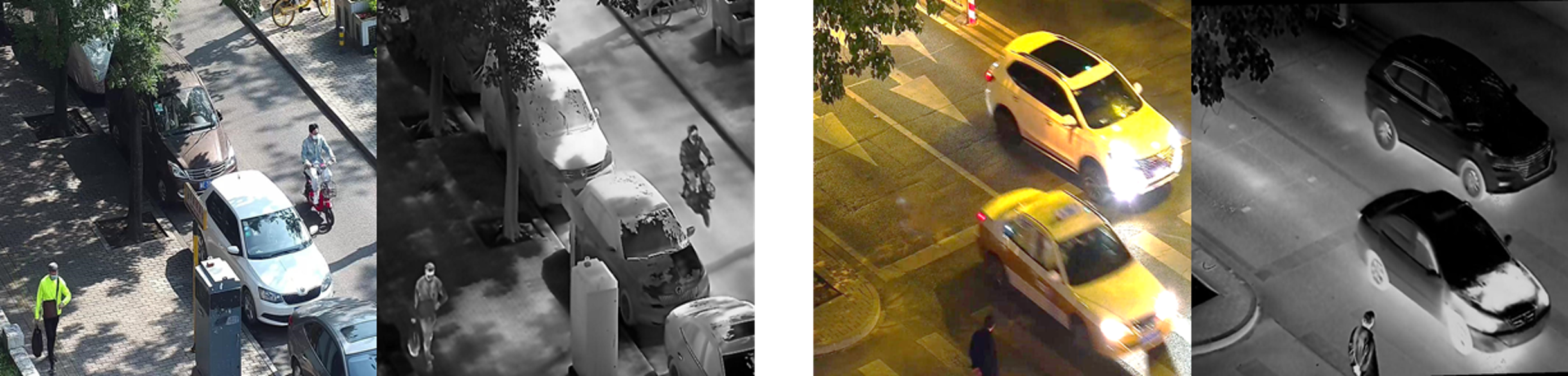}}
	\vspace{8pt}
	\subfloat[DroneVehicle (Optical-Infrared) \label{fig:dataset_dronevehicle}]{\includegraphics[width=1\linewidth]{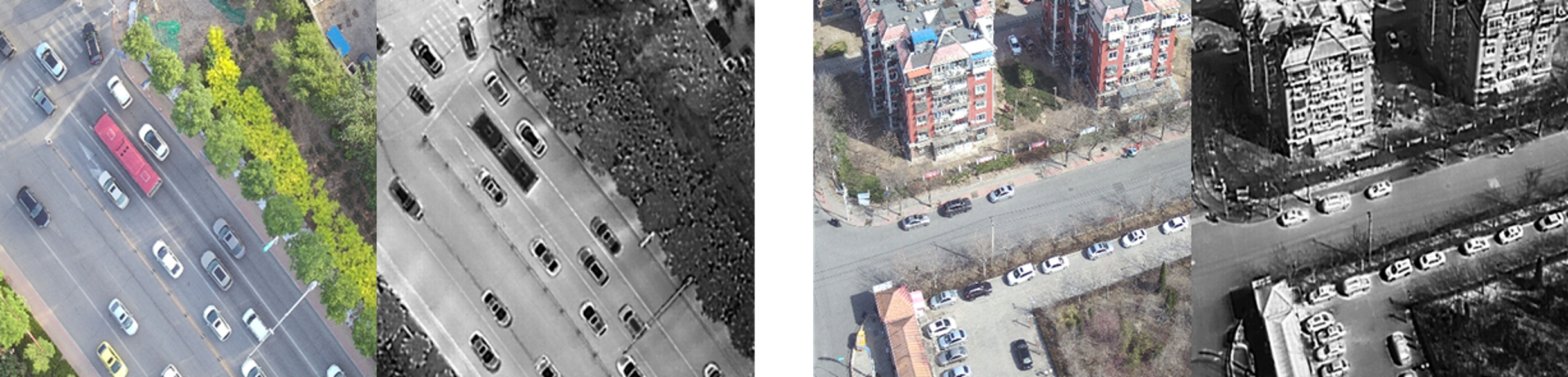}}
	\vspace{8pt}
		\subfloat[OSdataset (Optical-SAR) \label{fig:dataset_osdataset}]{\includegraphics[width=1\linewidth]{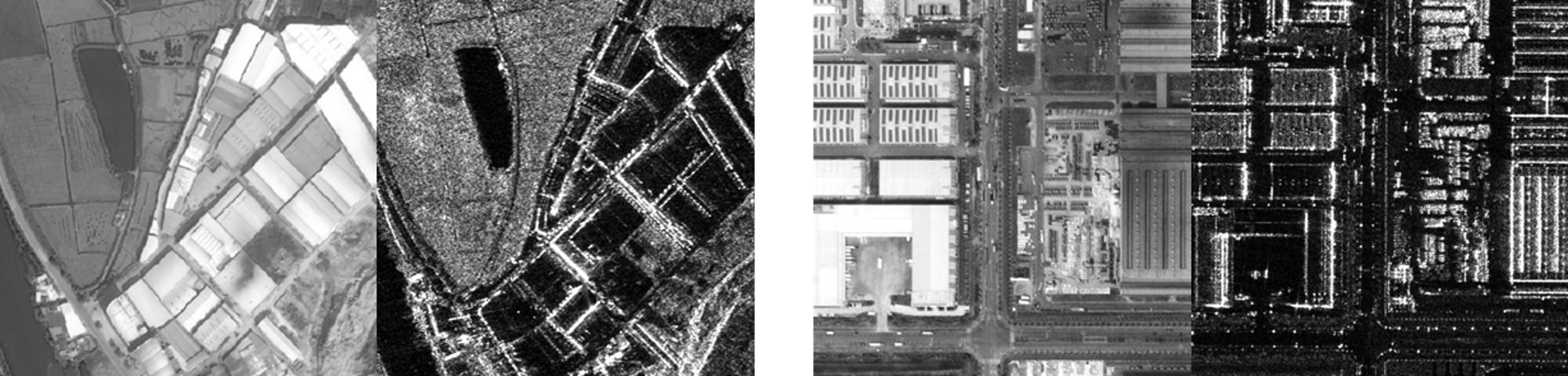}}
	\caption{Examples of Multimodal Images from Experimental Datasets}
	\label{fig:dataset}
	\vspace{-15pt}
\end{figure}

\textbf{Datasets.} We adopt three public datasets for the main comparative experiments and analyses: LLVIP \cite{jia2021llvip}, DroneVehicle \cite{sun2022drone}, and OSdataset \cite{xiang2020automatic}. In addition, we use the public Multimodal Image Matching Dataset (MIMD) \cite{jiang2021review} for generalization analysis.

(1) The LLVIP dataset comprises 15,488 pairs of close-range visible-infrared images, primarily featuring ground objects such as vehicles, pedestrians, roads, and trees, as shown in Fig. \ref{fig:dataset_llvip}. Visible images contain rich texture, but targets can be difficult to recognize under poor illumination; in contrast, thermal infrared images capture scenes based on the surface temperature field of objects, allowing them to highlight key targets such as pedestrians and vehicles, yet they lack texture information. Consequently, the LLVIP dataset exhibits significant radiometric differences, posing a challenge for multimodal image matching methods. We follow the official dataset split, using 12,025 image pairs as the training set, and 3,463 as the test set.

(2) The DroneVehicle dataset has a resolution of approximately 0.2 m and contains 28,439 pairs of optical-infrared images acquired by drones, covering urban roads, vehicles, parking lots, roadside buildings, and intersections, as shown in Fig. \ref{fig:dataset_dronevehicle}. We follow the official dataset split, using 17,990 image pairs as the training set, 1,469 as the validation set, and 8,980 as the test set.

(3) The OSdataset has a resolution of 1 m and comprises 2,673 pairs of optical-SAR images. The SAR images were acquired by the multipolarized C-band Gaofen-3 satellite, and the optical images were downloaded from Google Earth Engine, covering scenes such as cities, airports, rivers, farmlands, lakes, and highways, as shown in Fig. \ref{fig:dataset_osdataset}. Optical-SAR image matching is one of the most difficult tasks in multimodal image matching, because of irrelevant texture details, local geometric distortions, significant NRD between images, blurred SAR edges, and severe speckle noise. We follow the official dataset split, using 2011 image pairs as the training set, 238 as the validation set, and 424 as the test set.

(4) The MIMD dataset, collected from remote sensing, medical, and computer vision domains, contains 18 distinct multimodal pair types (e.g., day-night, image-map, optical-LiDAR, cross-season, RGB-NIR, T1-T2, and health-disease). The dataset is not used for training. It is therefore well-suited for the generalization experiments in Section \ref{sec:generalization}, which evaluate whether the RRSI method can maintain matching performance when applied to unseen cross-modal data without additional training on such modalities.

\textbf{Benchmark Settings.} The proposed RRSI method is compared with nine state-of-the-art methods for a comprehensive analysis of matching performance, including the handcrafted feature-based methods RIFT \cite{li2019rift} and WSSF \cite{wan2023multimodal}, the detector-based learning methods LightGlue \cite{lindenberger2023lightglue}, $\mathrm{MINIMA}_\mathrm{LightGlue}$ \cite{ren2025minima}, and SceneGlue \cite{du2026sceneglue}, as well as the detector-free learning methods LoFTR \cite{sun2021loftr}, EfficientLoFTR \cite{wang2024efficient}, XoFTR \cite{tuzcuouglu2024xoftr}, and JamMa \cite{lu2025jamma}.

\textbf{Evaluation Metrics.} We employ the following metrics to evaluate image matching methods:

(1) Root Mean Square Error ($\mathrm{RMSE}$): For each test pair, DEGENSAC estimates a homography $\hat{H}$ from the matched correspondences, while the predefined test transformation parameters provide the ground-truth homography $H_{\mathrm{gt}}$. The four corner points of the source image are projected using both $\hat{H}$ and $H_{\mathrm{gt}}$, and the RMSE is computed from the pixel-coordinate differences between the two sets of projected corners, thereby measuring the deviation of the estimated registration from the ground-truth transformation. For every test pair, the RMSE is capped at 10 pixels. If the computed RMSE exceeds 10 pixels, or if matching or registration fails because of insufficient correspondences, homography estimation failure, or any other failure condition, the RMSE of that pair is set to 10 pixels. The reported RMSE is the mean of these capped per-pair values over the entire test set.

(2) Success Rate ($\mathrm{SR}$): The ratio of the number of successfully registered image pairs to the total number of image pairs in the test dataset. An image pair is considered successfully registered if its $\mathrm{RMSE}$ is below a reprojection error threshold $T$ (in pixels).

(3) Area Under the Curve ($\mathrm{AUC}$): Defined as the normalized area under the curve of $\mathrm{SR}$ plotted against the localization error threshold $T$. It is used to quantify the overall performance of an image matching algorithm under different error tolerances. Specifically, the $\mathrm{SR}-T$ curve is constructed using piecewise-linear interpolation, and the integral area under the curve over the interval $[0, T_0]$ is calculated using the trapezoidal rule. The integral result is then divided by $T_0$ to normalize the AUC to the range [0, 1], where $T_0$ typically takes values such as 3, 5, or 10 pixels. The mathematical expression is:

\begin{equation}
	\mathrm{AUC}(T_0)=\frac{1}{T_0} \int_{0}^{T_0} \mathrm{SR}(t)dt
\end{equation}

\textbf{Experimental details.} (1) All deep learning-based comparison methods are initialized from the official pre-trained models released by their authors and then transfer-trained on the corresponding training datasets used in this study, following the official configurations and recommended training settings wherever applicable. (2) Regarding keypoint detector configurations: RIFT and WSSF use their built-in traditional keypoint detectors, while LightGlue, $\mathrm{MINIMA}_\mathrm{LightGlue}$, and SceneGlue use the pre-trained SuperPoint detector. The maximum number of keypoints per image is set to 5000 for detector-based methods where applicable, consistent with the RRSI method. LoFTR, EfficientLoFTR, XoFTR, and JamMa do not require an external keypoint detector. (3) For accuracy assessment, matching points of all methods are processed using the DEGENSAC \cite{chum2005two} outlier rejection algorithm, with a homography matrix adopted as the fitted geometric transformation model. (4) For evaluations involving randomly generated rotations, scale changes, and other geometric perturbations, the transformation parameters and corresponding ground-truth homographies are generated once for every test pair and saved before evaluation. All comparison methods use exactly the same transformed images and parameter files. For example, the 424 test pairs of the OSdataset share one fixed protocol file containing 424 sets of transformation parameters. (5) All experiments were conducted on a hardware platform equipped with dual Intel Xeon Platinum 6326 CPUs (2.90 GHz base frequency, 16 cores, 32 threads, 24 MB L3 cache, 185 W TDP) and eight NVIDIA GeForce RTX 4090 GPUs (each with 24 GB VRAM).
\begin{table*}
	\caption{Quantitative Results of Image Matching on the LLVIP, DroneVehicle, and OSdataset (\textbf{Bold} Indicates the Optimal Value, and \underline{Underline} Indicates the Sub-optimal Value)}
	\centering
	\begin{tabular}{ccccccccccccc}
		\toprule
			\multirow{3}{*}{Methods } & \multicolumn{4}{c}{LLVIP} &  \multicolumn{4}{c}{DroneVehicle} & \multicolumn{4}{c}{OSdataset} \\
		\cline{2-5}
		\cline{6-9}
		\cline{10-13}
		& RMSE & \multicolumn{3}{c}{AUC/(\%)} &RMSE & \multicolumn{3}{c}{AUC/(\%)} & RMSE & \multicolumn{3}{c}{AUC/(\%)} \\
		& /px       & @3px & @5px & @10px & /px       & @3px & @5px & @10px & /px       & @3px & @5px & @10px \\
		\midrule
		{RIFT}  \cite{li2019rift}
		& {5.01} & {3.27} & {9.38} & {24.14}                
		& {5.04} & {3.46} & {8.19} & {19.37}   
		& 4.74 & 4.05   & 9.15 & 20.99\\
		
		{WSSF} \cite{wan2023multimodal}
		&{5.36} &{2.87} &{7.42} & {21.39}                
		& {5.21} &{4.74} &{9.52} & {21.30}    
		& 5.09 & 3.81   & 10.54 & 26.57\\  
		LoFTR \cite{sun2021loftr}
		& 3.17 & {18.65} & {39.99} & {67.03}
		& 3.40 & {20.66} & {37.23} & {59.99}
		& {3.72} & {11.12} & {29.06} &{55.40}\\
		
		{LightGlue} \cite{lindenberger2023lightglue}
		& {3.37} & {17.27} & {36.99} & {66.36}
		& {2.86} & {25.08} & {45.27} & {69.02}
		& 3.53  & 14.07 & 35.26 & 64.53\\ 
		
		{EfficientLoFTR}  \cite{wang2024efficient}
		& 3.14 & 19.92 & 39.93 & 68.42 
		& {2.94} & {24.72} & {44.66} &{67.05}  
		& 3.44 & 14.39 & 35.73 & 64.57\\  
		
		{XoFTR} \cite{tuzcuouglu2024xoftr}
		&{2.91} & \underline{23.67} & {44.18} &{70.49}
		& {3.42}& {17.63} &{36.74} &{62.34}
		& {3.24} & {16.71} & {38.07}  & 64.92\\ 
		
		$\mathrm{MINIMA}_\mathrm{LightGlue}$ \cite{ren2025minima}                                                   
		&{2.89}&20.24&{43.63}&{71.19}   
		&{2.74}&{27.73}&{48.02}&{71.75} 
		& 3.34 &14.07 &36.37&{66.50}\\ 
				
		{JamMa} \cite{lu2025jamma} 
		& \underline{2.74} & {22.21} & \underline{46.59} &{71.85}
		& \underline{2.41} &  \textbf{37.40} & \textbf{53.82} & \underline{73.57}
		& {3.20} & \underline{17.37} & {39.34} &{66.12}\\
		
		SceneGlue \cite{du2026sceneglue}
		& 2.82 & 21.50 & 45.48 & \underline{72.28}               
		& 3.12 & 19.56 & 40.59 & 66.46          
		& \textbf{3.13} & 16.86  & \underline{39.93} & \underline{67.66}\\ 
		
		RRSI  
		& \textbf{2.70} & \textbf{25.48} & \textbf{47.35} & \textbf{73.10}      
		& \textbf{2.39} & \underline{33.05} & \underline{53.44} & \textbf{74.70}    
		& \underline{3.17} &  \textbf{17.97} & \textbf{40.05} & \textbf{67.83} \\  
		\bottomrule
	\end{tabular}
	
	\label{tab:exp_main}
\end{table*}

\begin{figure*}
	\centering
	\subfloat[LLVIP \label{fig:SR_llvip}]{\includegraphics[width=0.3\linewidth]{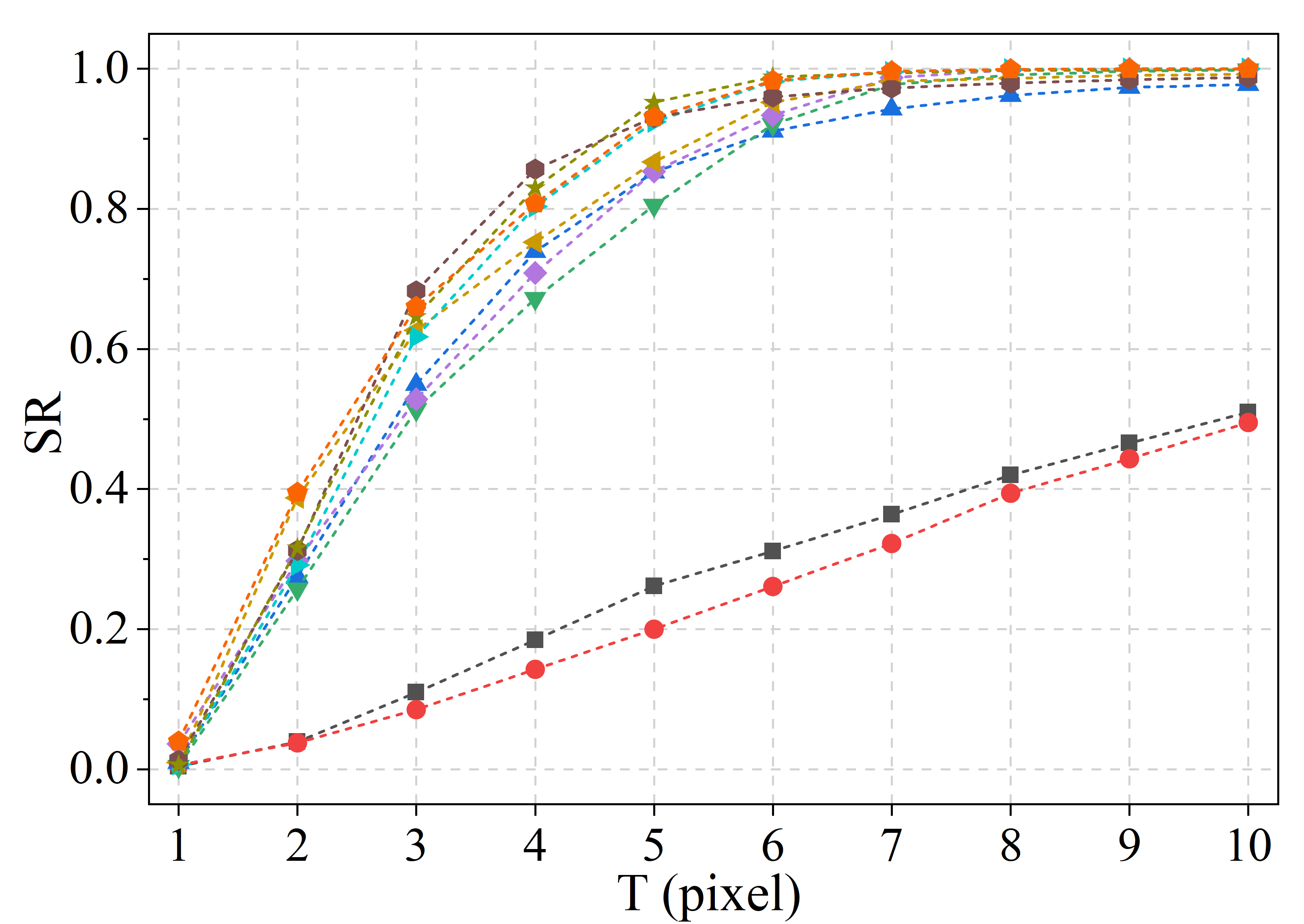}}
	\hfil
	\subfloat[DroneVehicle \label{fig:SR_dronevehicle}]{\includegraphics[width=0.3\linewidth]{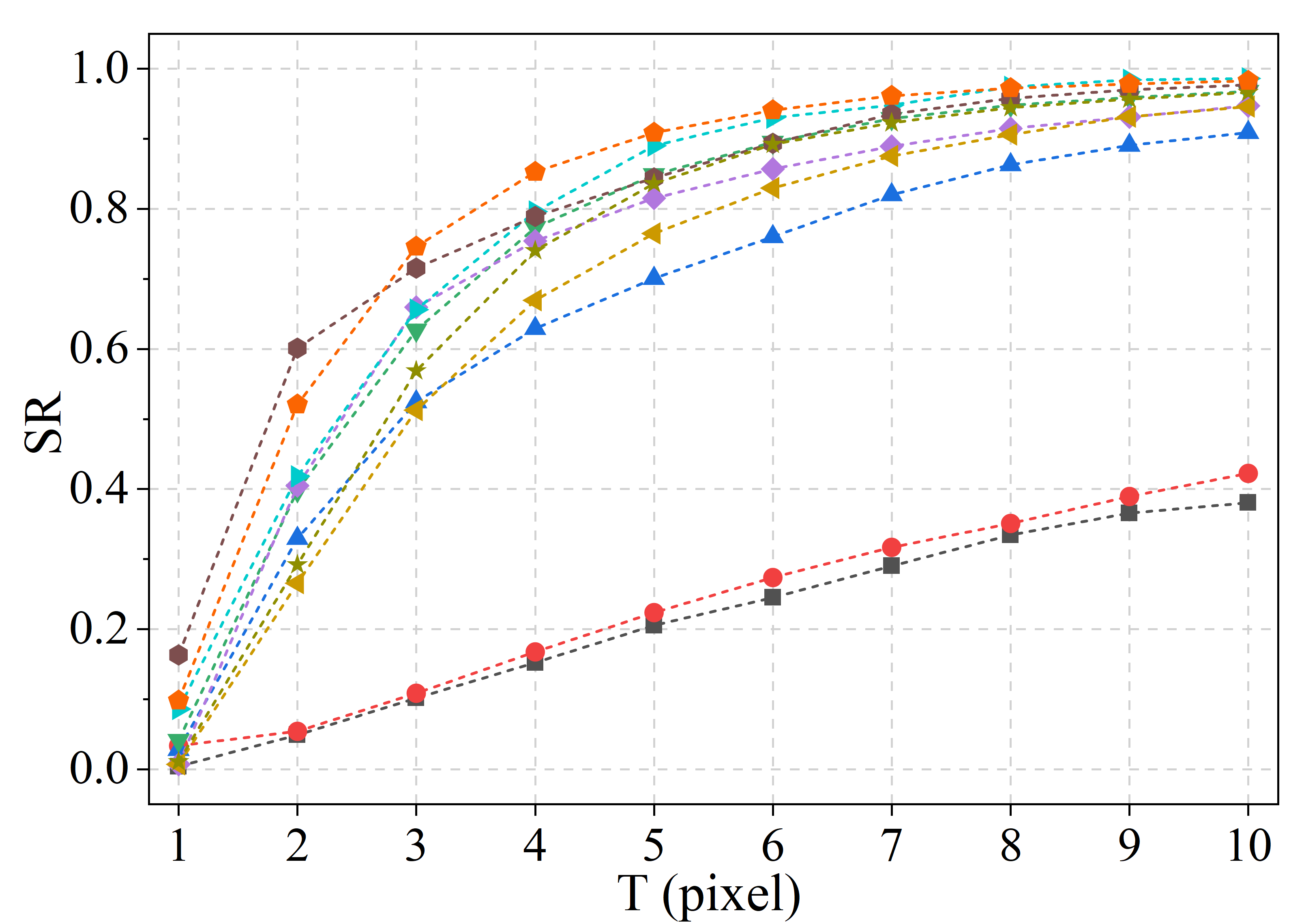}}
	\hfil
	\subfloat[OSdataset \label{fig:SR_osdataset}]{\includegraphics[width=0.3\linewidth]{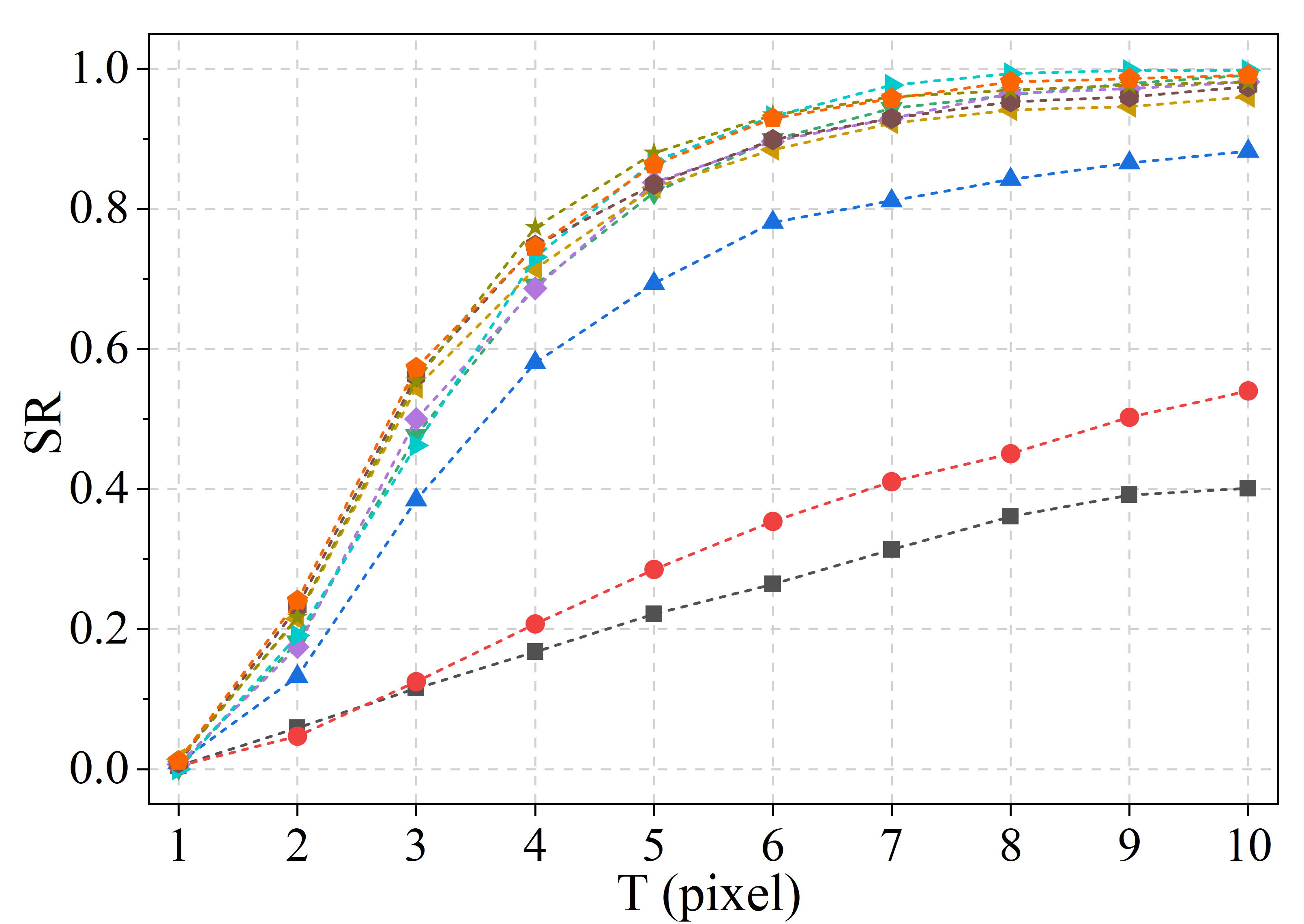}}
	\hfil
	\subfloat{\includegraphics[width=0.09579\linewidth]{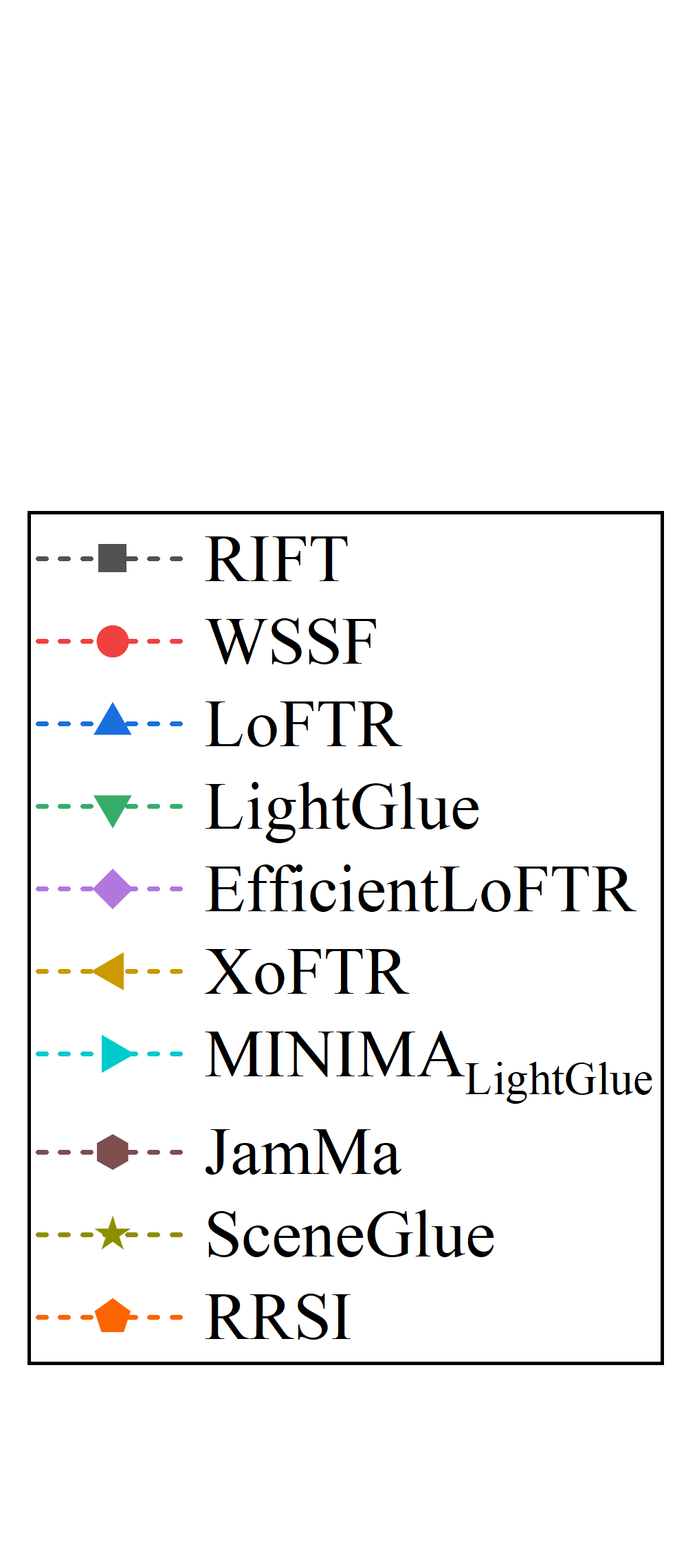}}
	\caption{SR-$T$ curve: image matching success rate at different reprojection error thresholds}
	\label{fig:SR}
	\vspace{-15pt}  %
\end{figure*}

\begin{figure*}[!t]
	\centering  %
	\subfloat[LLVIP (Visible-Infrared) \label{fig:visualize_llvip}]{\includegraphics[width=0.95\linewidth]{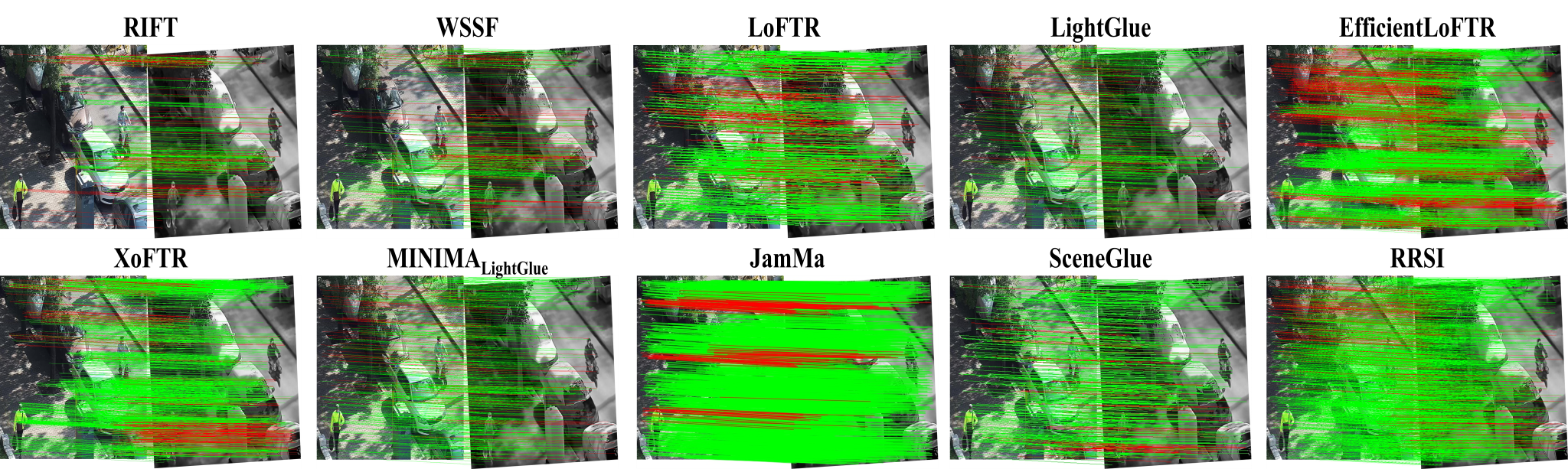}}
	\vspace{8pt}  
	\subfloat[DroneVehicle (Optical-Infrared) \label{fig:visualize_dronevehicle}]{\includegraphics[width=0.95\linewidth]{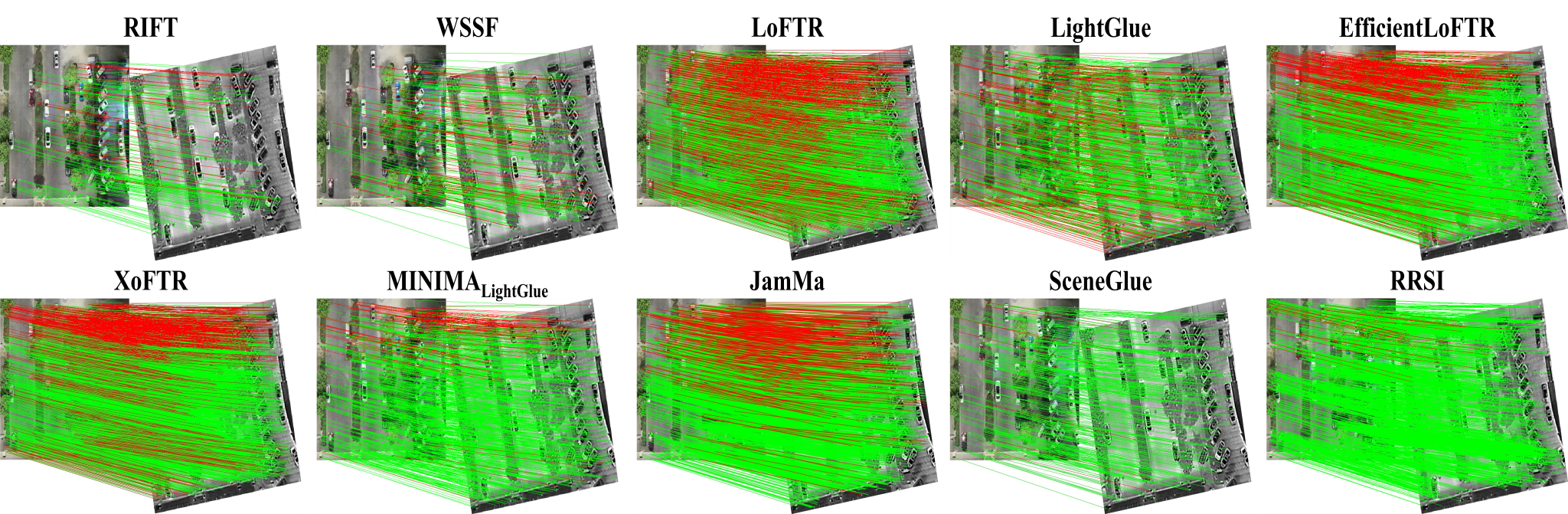}}
	\vspace{8pt}  
	\subfloat[OSdataset (Optical-SAR) \label{fig:visualize_osdataset}]{\includegraphics[width=0.95\linewidth]{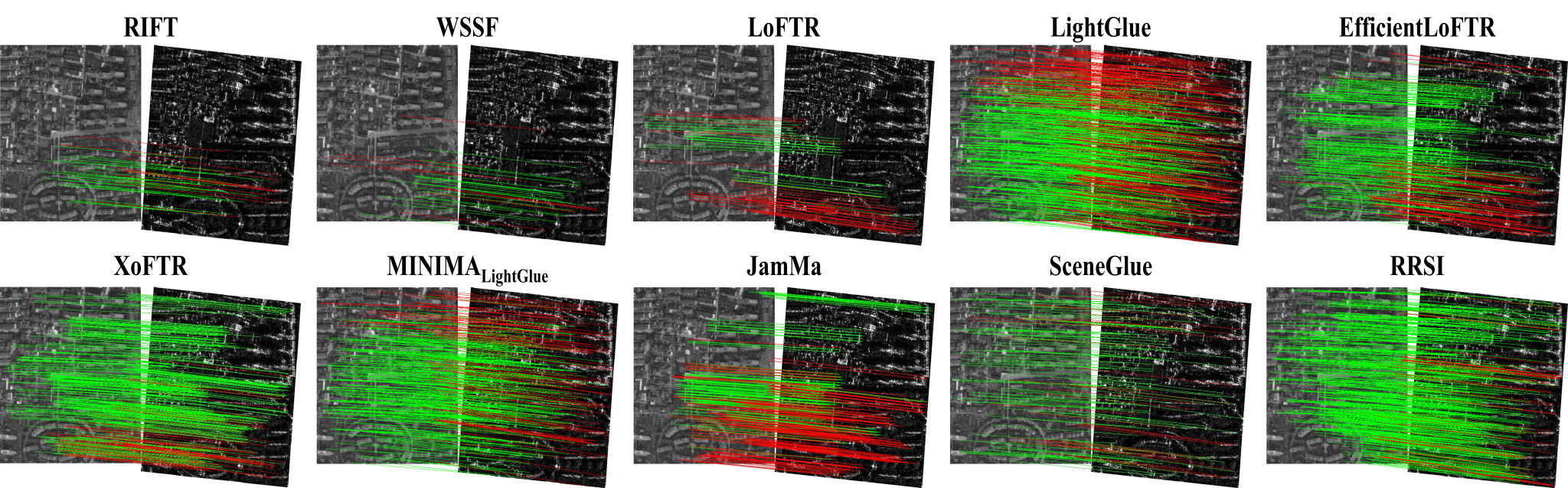}}
	\caption{Visualization of matching results by different methods}
	\label{fig:visualization}
	\vspace{-15pt}  %
\end{figure*}

\begin{figure*}[!b]
	\centering
	\subfloat[LLVIP \label{fig:Registration_llvip}]{\includegraphics[width=0.3\linewidth]{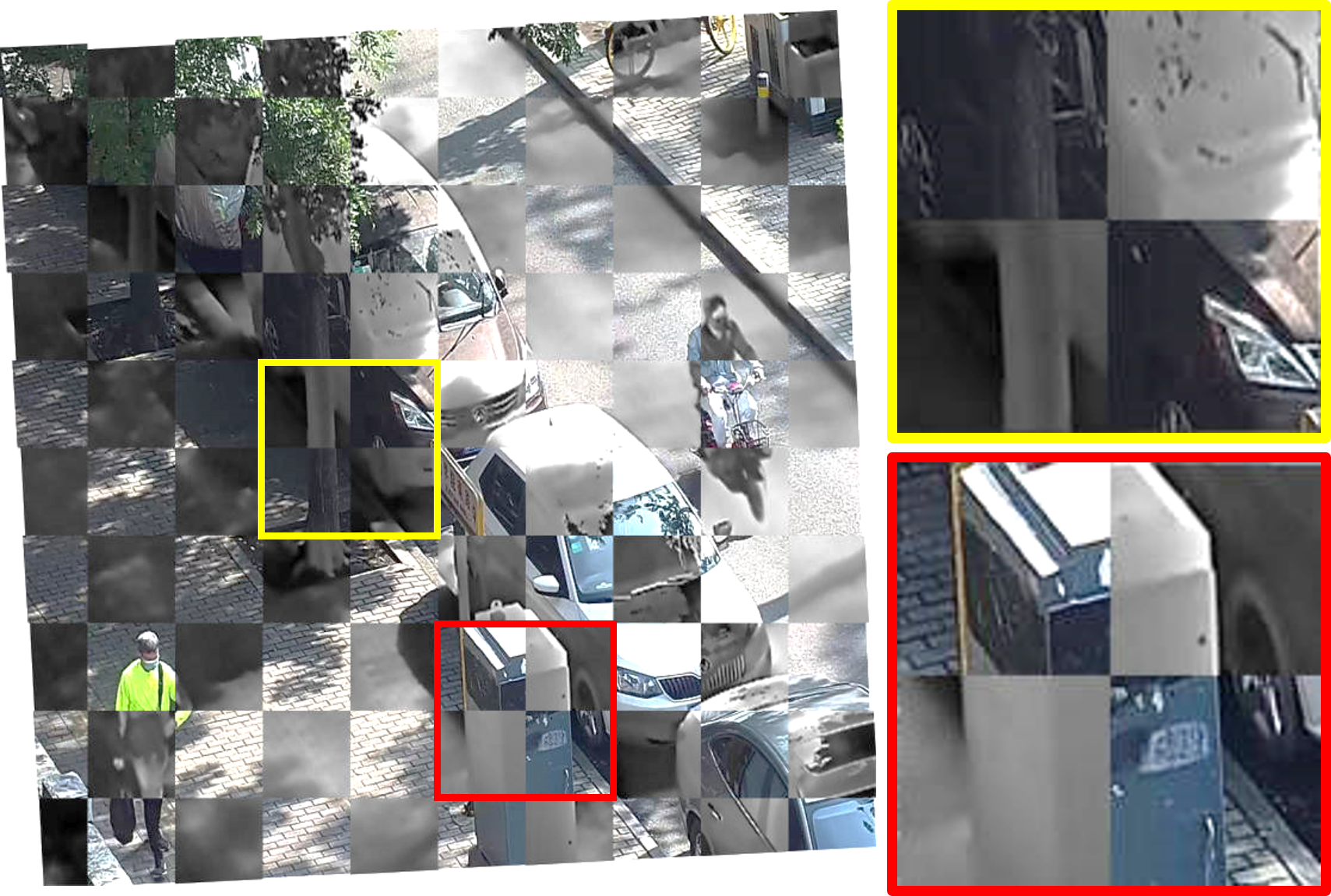}}
	\hfil
	\subfloat[DroneVehicle \label{fig:Registration_dronevehicle}]{\includegraphics[width=0.3\linewidth]{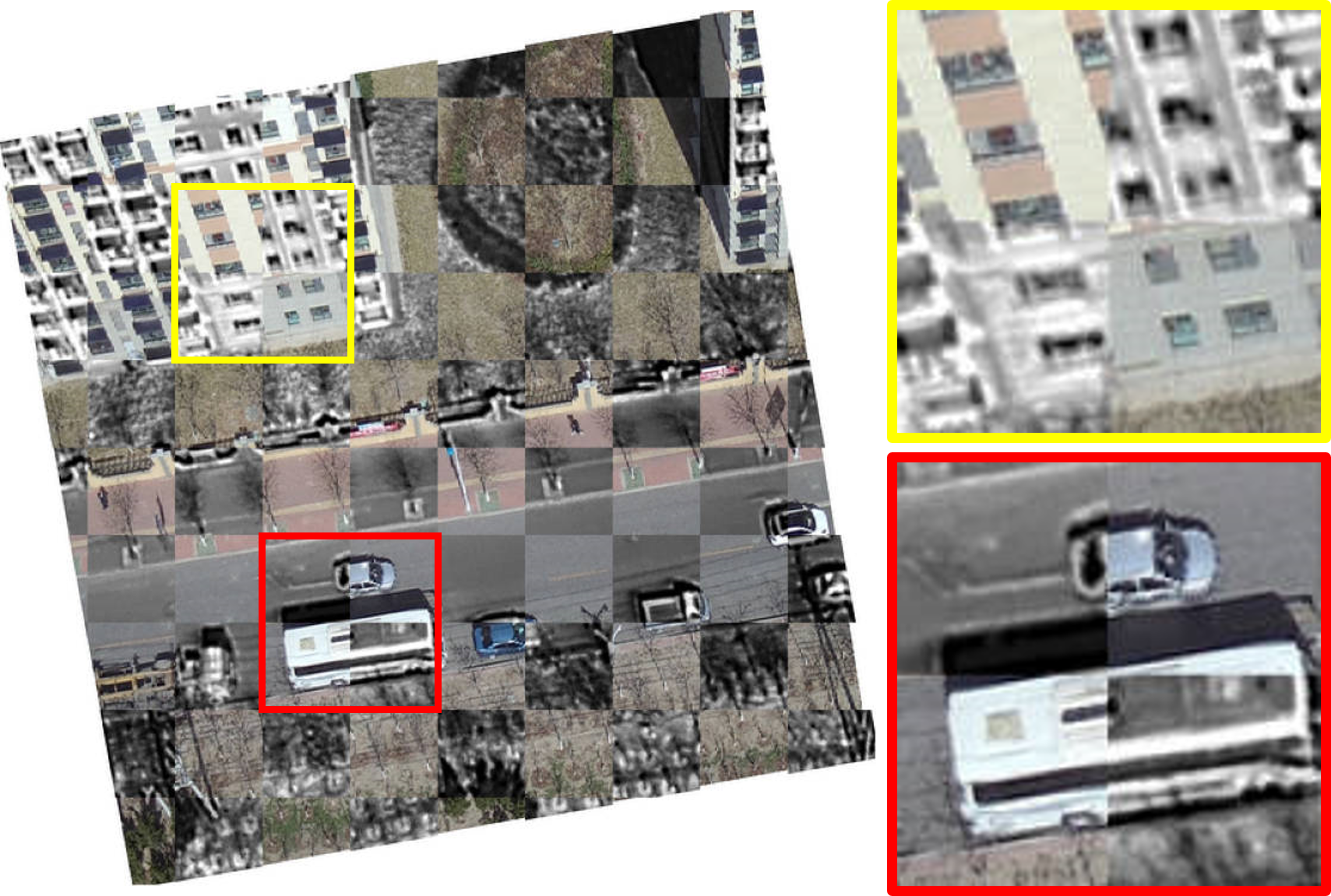}}
	\hfil
	\subfloat[OSdataset \label{fig:Registration_osdataset}]{\includegraphics[width=0.3\linewidth]{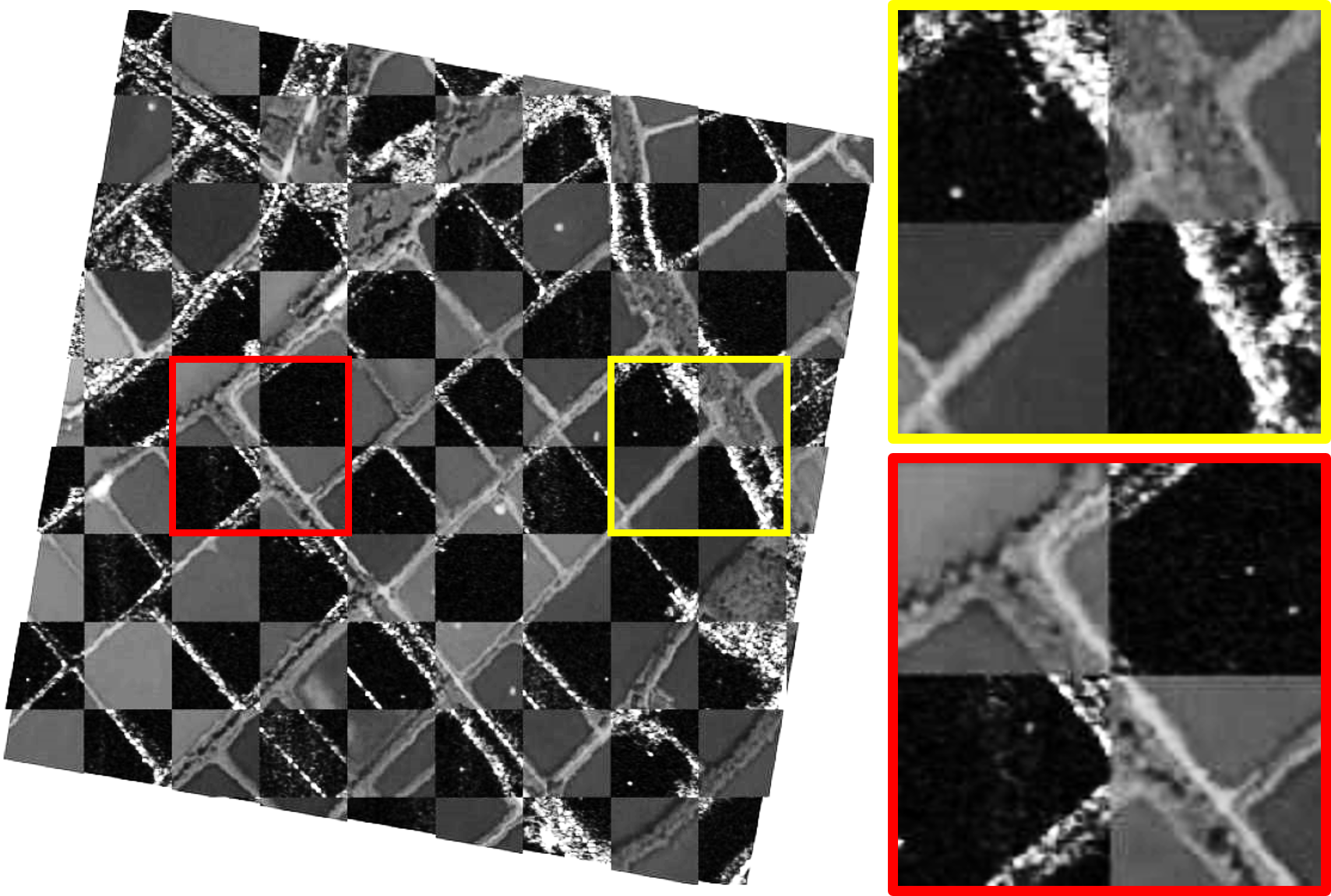}}
	\caption{Checkerboard visualization of the RRSI's registration results}
	\label{fig:Registration}
	\vspace{-15pt}  %
\end{figure*}

\subsection{Multimodal Image Matching Evaluation}
\label{sec:image matching}

\textbf{Quantitative Analysis.} The matching performance of the RRSI method and all comparative methods is quantitatively evaluated using RMSE, AUC, and the $\mathrm{SR}-T$ curve. Since the image pairs of the test dataset are pre-registered, we apply random rotations within the range of  $\pm$15°, random scaling within the range of $[1, 1.25]$, and perspective transformations within ±0.08 to the test dataset to simulate geometric distortions. Fig. \ref{fig:SR} shows the corresponding $\mathrm{SR}-T$ curves of various methods, and Table \ref{tab:exp_main} presents the RMSE as well as the AUC at the reprojection error thresholds of 3, 5, and 10 pixels. 

Because the datasets differ in modality, image resolution, scene content, and noise characteristics, the relative performance of the methods varies across datasets. On the LLVIP dataset with the visible-thermal infrared modality, targets in visible images are hard to distinguish under dark conditions, while thermal infrared images suffer from blurred texture details. This dataset effectively differentiates the performance of various matching methods. Among these, deep learning-based methods generally outperform traditional ones. Although XoFTR, $\mathrm{MINIMA}_\mathrm{LightGlue}$, JamMa, and SceneGlue provide competitive results, the RRSI achieves the best values across all reported metrics. Specifically, its RMSE is 2.70 pixels, and its AUC values under the 3/5/10-pixel thresholds are 25.48\%, 47.35\%, and 73.10\%, respectively. These results demonstrate that the RRSI has excellent adaptability to visible-thermal infrared image matching tasks under low-light conditions.

On the optical-infrared DroneVehicle dataset, the UAV images provide relatively clear textures and distinct structural details. Nevertheless, the traditional methods RIFT and WSSF yield RMSE values above 5 pixels and considerably lower AUC values than the learning-based methods, indicating limited robustness to the combined radiometric and geometric variations. Among the learning-based methods, JamMa achieves the highest AUC@3px and AUC@5px values of 37.40\% and 53.82\%, respectively, and obtains the second-lowest RMSE of 2.41 pixels. The RRSI follows closely at these strict thresholds, with AUC@3px and AUC@5px values of 33.05\% and 53.44\%, while achieving the lowest RMSE of 2.39 pixels and the highest AUC@10px of 74.70\%. These results demonstrate that the RRSI maintains a strong balance between strict localization accuracy and overall registration robustness on optical-infrared UAV imagery.

On the optical-SAR OSdataset, as nonlinear radiometric differences, local geometric distortions, and noise interference intensify, the $\mathrm{SR}-T$ curves of traditional methods (e.g., RIFT, WSSF) rise slowly, and their upper SR limits are significantly lower than those of deep learning-based methods. This reflects the high sensitivity of handcrafted features to data quality and cross-modal discrepancies. SceneGlue obtains the lowest RMSE of 3.13 pixels, followed closely by the RRSI at 3.17 pixels. More importantly, the RRSI achieves the highest AUC values of 17.97\%, 40.05\%, and 67.83\% under the 3/5/10-pixel thresholds, respectively, outperforming all comparison methods in overall success-rate performance. This demonstrates that the RRSI remains highly competitive under severe nonlinear radiometric differences, local geometric distortions, and SAR speckle noise.

\textbf{Qualitative Analysis.} Fig. \ref{fig:visualization} presents the visualization results of matched point correspondences (one example per dataset) for each method on the three datasets. For clarity, only the inliers retained after DEGENSAC fitting are displayed. Their reprojection errors are computed against the ground-truth homography derived from the predefined test transformation parameters, rather than the homography estimated by DEGENSAC. Green lines represent retained correspondences with a ground-truth reprojection error of less than 3 pixels, whereas red lines represent those with an error greater than or equal to 3 pixels. The denser the green lines and the lower the proportion of red lines, the smaller the overall reprojection error of the image and the more reliable the matching result. It can be observed that, in the visualization results across multiple datasets, the RRSI exhibits denser correct matching lines and a lower proportion of incorrect matching lines, which is consistent with the quantitative analysis results. Furthermore, Fig. \ref{fig:Registration} presents one registration result example from each of the three datasets using the RRSI method in a checkerboard format. The magnified regions demonstrate high registration accuracy. For instance, edges of targets across different modalities connect smoothly: these include tree trunks in Fig. \ref{fig:Registration_llvip}, vehicles and buildings in Fig. \ref{fig:Registration_dronevehicle}, and roads and farmlands in Fig. \ref{fig:Registration_osdataset}. Only limited visible misalignment remains, further demonstrating the strong matching performance of RRSI.

\begin{figure}
	\centering
	\includegraphics[width=0.8\columnwidth]{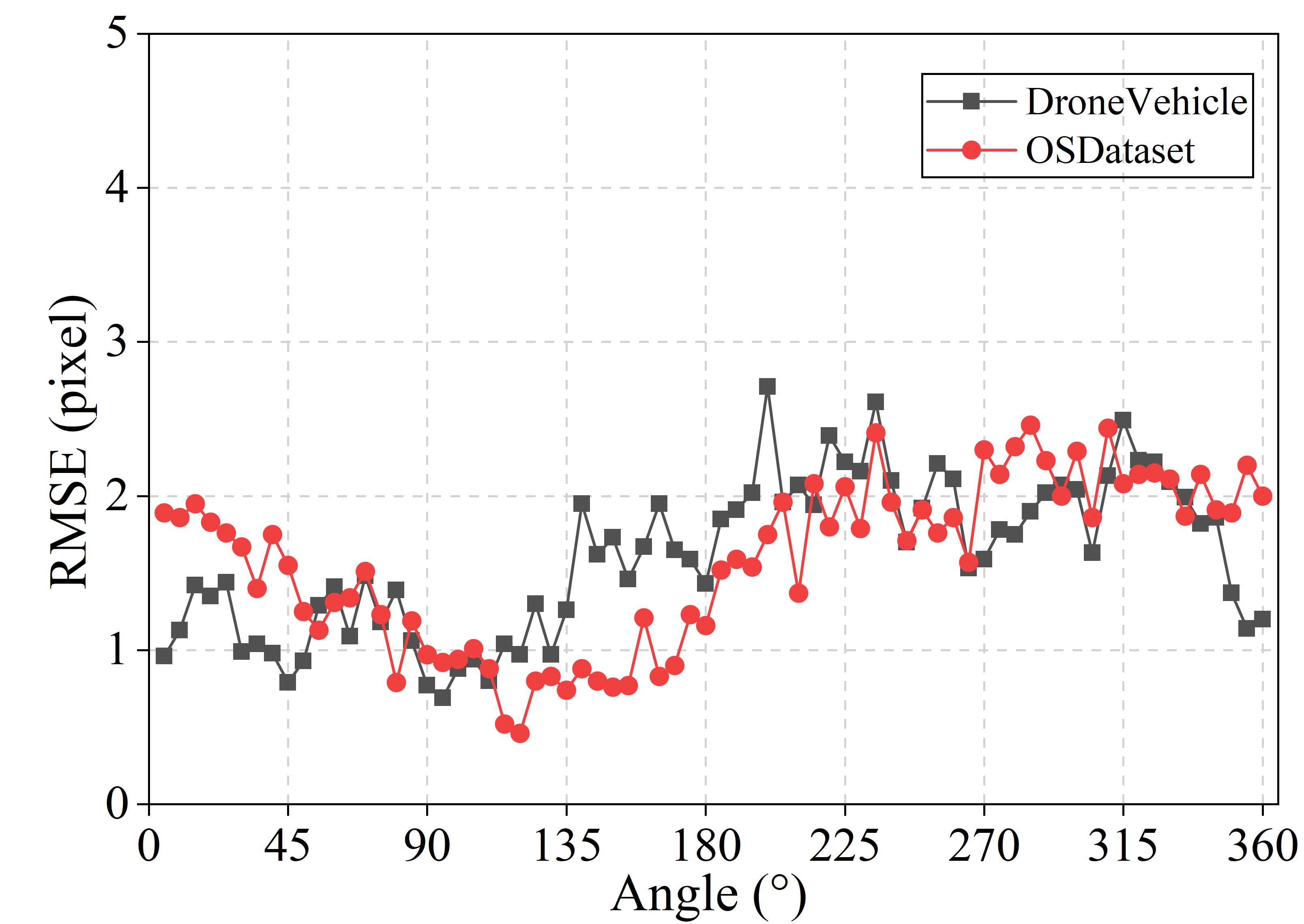} 
	\caption{RMSE under different rotational angles} %
	\label{fig:rot_inv} 
	\vspace{-5pt}
\end{figure}

\begin{figure}
	\centering  %
	\subfloat[DroneVehicle \label{fig:rot_inv_dronevehicle}]{\includegraphics[width=1\linewidth]{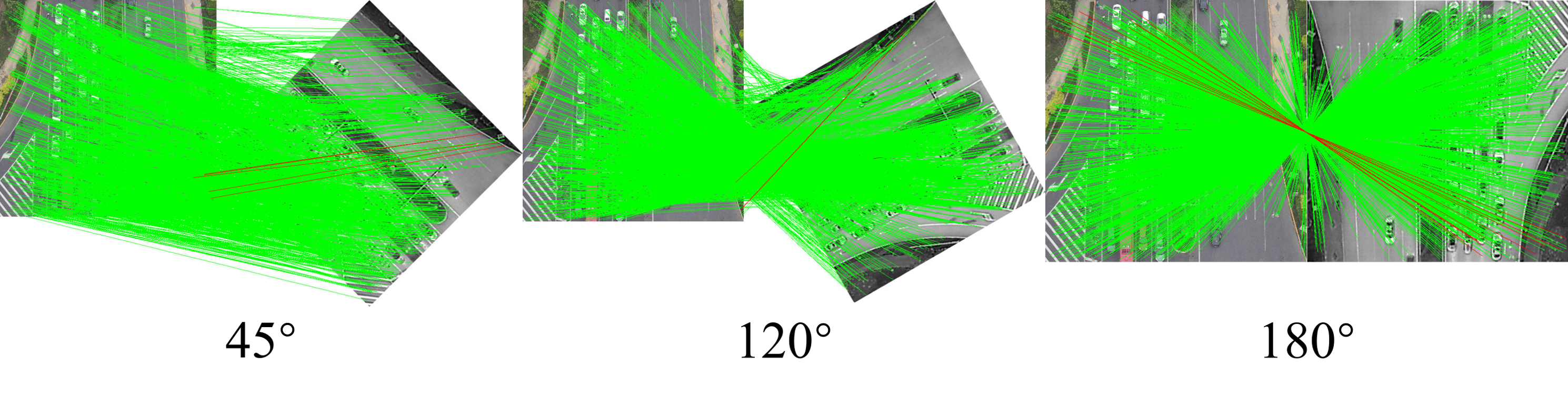}}
	\vspace{8pt}  %
	\subfloat[OSdataset \label{fig:rot_inv_osdataset}]{\includegraphics[width=1\linewidth]{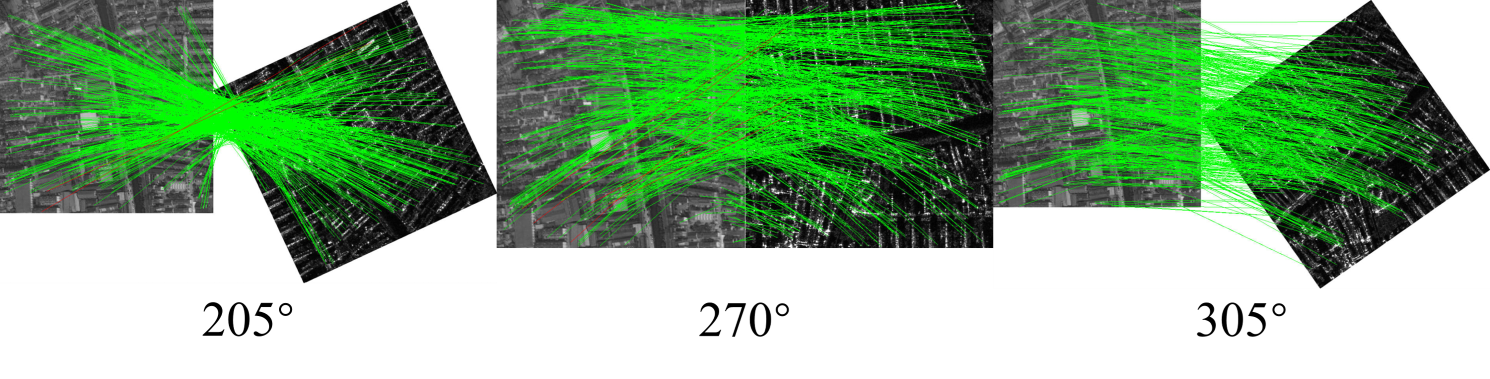}}
	\caption{Visualization of matching results under different rotational angles}
	\label{fig:rot_inv_visualize}
	\vspace{-15pt}  %
\end{figure}

\subsection{Rotation Invariance Analysis}
\label{sec:rotate invariance}

Rotation invariance is a crucial property of the RRSI. This section evaluates the adaptability of the RRSI to robust matching under rotation angles in the range $[0^\circ,360^\circ]$. Specifically, one pair of multimodal images is selected from each of DroneVehicle and OSdataset: optical images are used as reference images, while their corresponding infrared or SAR images are rotated counterclockwise around their center points. For each reference image, 72 rotated variants are generated at 5° increments (0°, 5°, $\ldots$, 355°), covering a full 360° rotation cycle. After performing matching for each of these images, the RMSE values are calculated and recorded in the line chart of Fig. \ref{fig:rot_inv}. It can be observed that on both datasets, the matching reprojection RMSE fluctuates within the range of approximately 1 to 2 pixels under most rotation angles. Neither dataset exhibits an abnormal increase in RMSE, and all results satisfy the criterion for correct matching. Additionally, Fig. \ref{fig:rot_inv_visualize} presents the visualized matching results of the RRSI method on multimodal images with rotational differences. It can be seen that the matched points are densely and evenly distributed under different rotation angles, with few mismatched points. This demonstrates that the RRSI method maintains robust matching across the full 360° rotation cycle, while also resisting radiometric differences.

\subsection{Scale Invariance Analysis}
\label{sec:scale invariance}
\begin{figure}
	\centering
	\includegraphics[width=0.8\columnwidth]{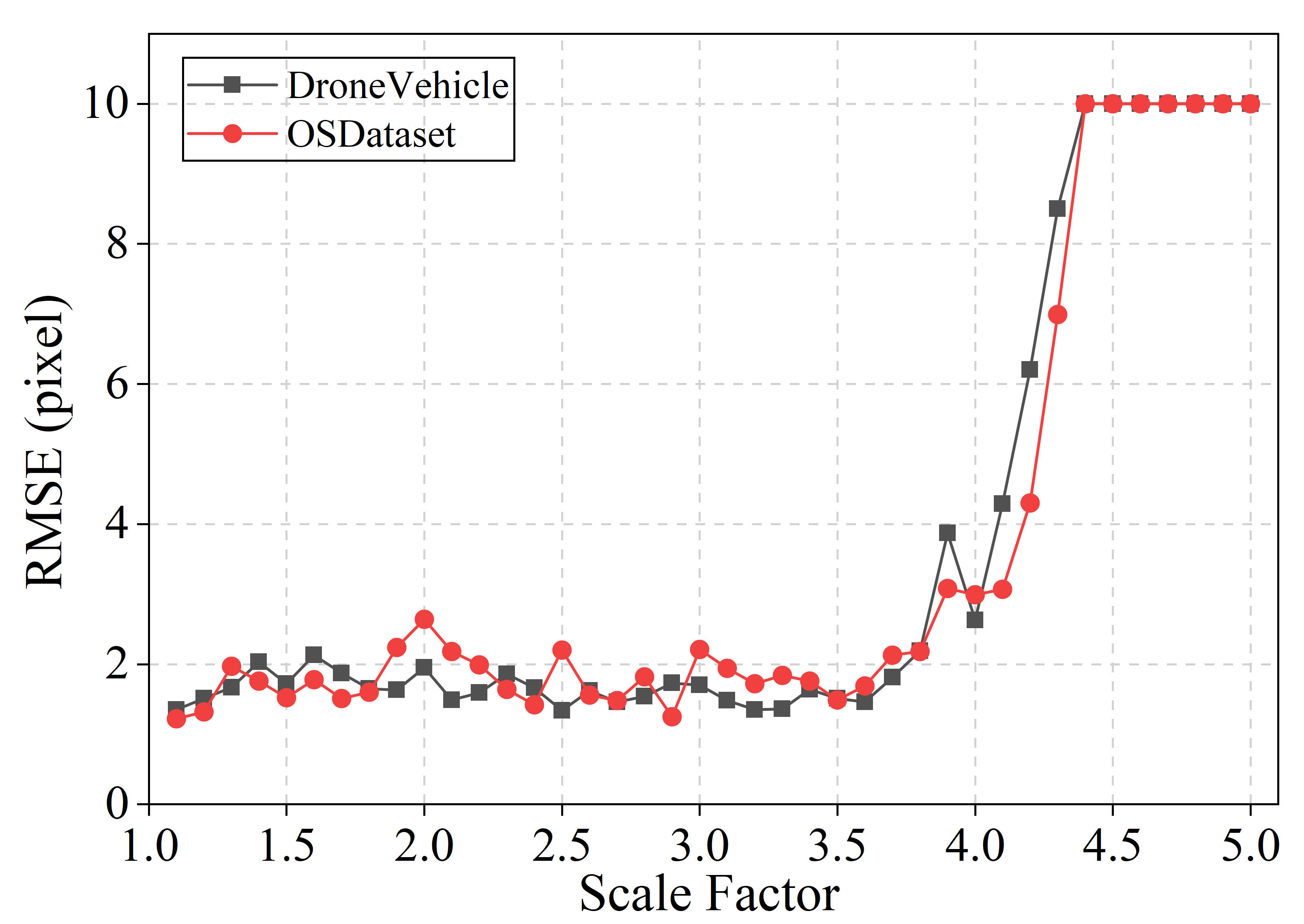} 
	\caption{RMSE under different scale factors, where values exceeding 10 pixels are clipped to 10 pixels} %
	\label{fig:sca_inv} 
	\vspace{-5pt}
\end{figure}

\begin{figure}
	\centering  %
	\subfloat[DroneVehicle \label{fig:sca_inv_dronevehicle}]{\includegraphics[width=1\linewidth]{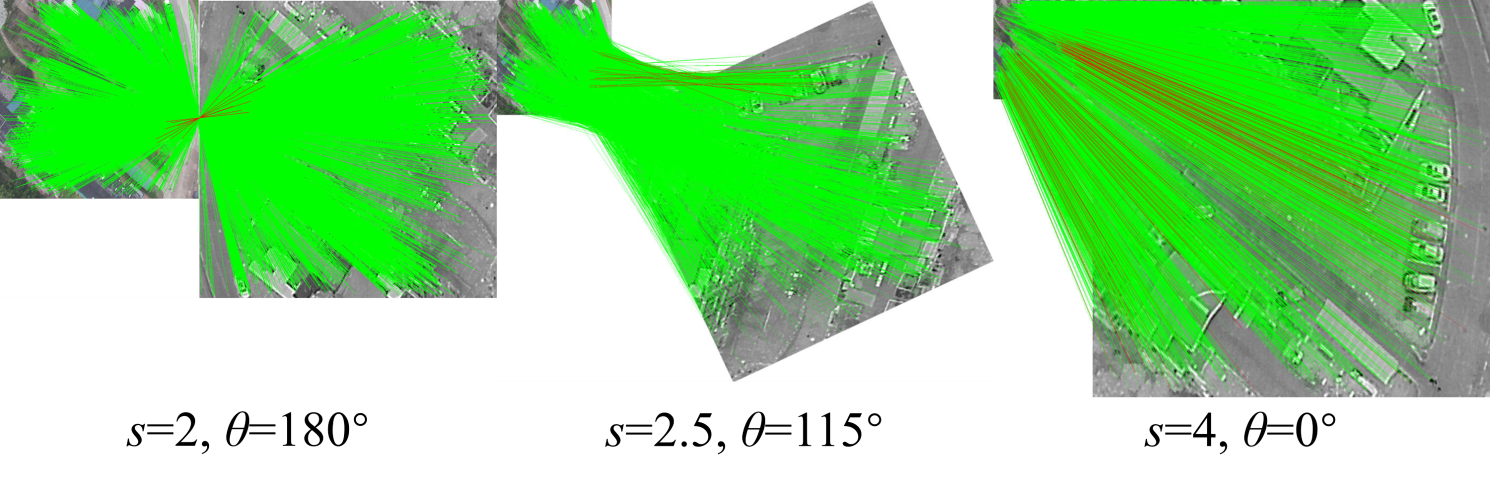}}
	\vspace{8pt}  %
	\subfloat[OSdataset \label{fig:sca_inv_osdataset}]{\includegraphics[width=1\linewidth]{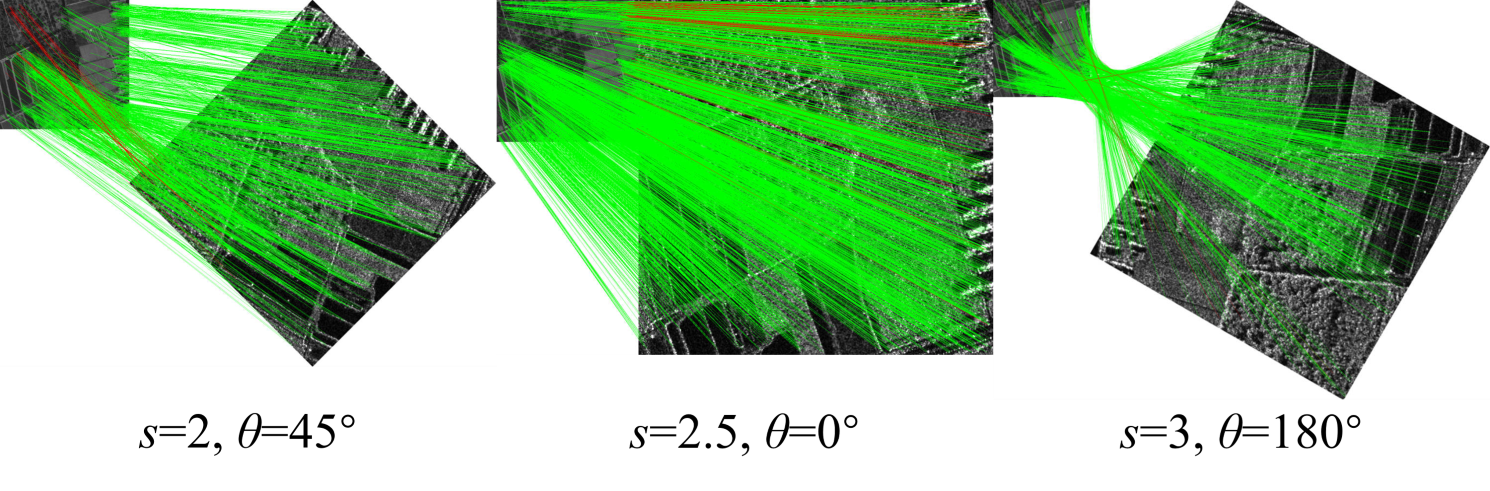}}
	\caption{Visualization of matching results under different scale factors $s$ and rotational angles $\theta$}
	\label{fig:sca_inv_visualize}
	\vspace{-15pt}  %
\end{figure}

Scale invariance is another key property of the RRSI. This section evaluates the matching adaptability of RRSI under image scale variations. Specifically, one pair of multimodal images is selected from each of DroneVehicle and OSdataset: optical images serve as reference images, while scale factors ranging from 1.1 to 5.0 at intervals of 0.1 are applied to their corresponding infrared or SAR images. For each reference image, a total of 40 images to be matched are obtained. After performing matching for each of these images, the RMSE values are calculated and recorded in the line chart of Fig. \ref{fig:sca_inv}. It can be observed that across both datasets: when the scale factor is in the range of [1.1, 3.7], the matching reprojection RMSE fluctuates within approximately 2 pixels; for scale factors in the range of [3.7, 4.1], the matching reprojection RMSE begins to increase, fluctuating within approximately $[2, 4]$ pixels; when the scale factor is in the range of [4.1, 4.4], the matching reprojection RMSE rises significantly but remains within 10 pixels; and when the scale factor exceeds 4.4, the matching results become unreliable. Additionally, Fig. \ref{fig:sca_inv_visualize} presents the visualized matching results of the RRSI method on multimodal images with scale factors and rotational angles. It can be seen that the matched points are densely and evenly distributed under different scale factors, with few mismatched points. In summary, we conclude that the proposed RRSI method can reliably match multimodal images with scale factors of up to four.

\subsection{Ablation Study}
\label{sec:ablation study}

\textbf{Effectiveness of the DHRS module.} This ablation experiment is conducted on the OSdataset. For the local region sampling step, the settings are as follows: (1) Cartesian Only: Performs only Cartesian sampling, removes the Log-Polar sampling and the subsequent intra-modal attention, replaces \(F'''_{A} / F'''_{B}\) with \(F'_{A,C} / F'_{B,C}\) in Fig. \ref{fig:framework}, and is followed by the inter-modal attention. (2) Log-Polar Only: Performs only Log-Polar sampling, removes the Cartesian sampling and the subsequent intra-modal attention, replaces \(F'''_{A} / F'''_{B}\) with \(F'_{A,LP} / F'_{B,LP}\) in Fig. \ref{fig:framework}, and is followed by the inter-modal attention. (3) Cartesian \& Log-Polar: This refers to the complete RRSI method based on the DHRS module.

The experimental results, as shown in Table \ref{tab:ablation1}, validate the effectiveness of the DHRS module. Specifically, Cartesian sampling retains the spatial structures of local regions, while Log-Polar sampling enhances the robustness of descriptors to scale and rotation variations. The DHRS module performs both Cartesian and Log-Polar sampling simultaneously, acquires dual patches, and is followed by parallel encoding—this synergy enhances the model’s matching performance in complex multimodal scenarios.

\begin{table}
	\caption{Ablation Results of the DHRS Module on the OSdataset (Optical-SAR)}
	\centering
	\begin{tabular}{ccccc}
		\toprule
		\multirow{2}{*}{Ablation Setting} & \multirow{2}{*}{RMSE/px} & \multicolumn{3}{c}{AUC/(\%)}         \\ 
		\cline{3-5}
		& & @3px & @5px & @10px\\
		\midrule
		Cartesian Only & 3.60 & 16.51 & 37.82 & 64.81 \\ 
		Log-Polar Only & 3.72 & 15.34 & 36.19 & 62.97 \\ 
		Cartesian \& Log-Polar & \multirow{2}{*}{3.17} & \multirow{2}{*}{17.97} &\multirow{2}{*}{40.05} &  \multirow{2}{*}{67.83}\\
		(the complete RRSI) \\ 
		\bottomrule
	\end{tabular}
	\label{tab:ablation1}
\end{table}

\begin{table}
	\caption{Ablation Results for the Joint Encoding of Geometric and Radiometric Relations on the OSdataset (Optical-SAR)}
	\centering
	\begin{tabular}{ccccc}
		\toprule
		\multirow{2}{*}{Ablation Setting} & RMSE & \multicolumn{3}{c}{AUC/(\%)}         \\ 
		\cline{3-5}
		&/px & @3px & @5px & @10px\\
		\midrule
		Baseline & 3.48 & 16.87 & 38.27 & 65.36\\ 
		Intra-Mod. Att. & 3.41 & 17.42 & 39.07 & 66.26 \\ 
		Inter-Mod. Att. & 3.25 & 17.74 & 39.45 & 66.90 \\ 
		Intra \& Inter-Mod. Att. & \multirow{2}{*}{3.17} & \multirow{2}{*}{17.97} &\multirow{2}{*}{40.05} &  \multirow{2}{*}{67.83}\\
		(the complete RRSI) \\ 
		\bottomrule
	\end{tabular}
	\label{tab:ablation2}
\end{table}

\begin{table}
	\caption{Ablation Results of the Generative Reconstruction Constraint on the OSdataset (Optical-SAR)}
	\centering
	\begin{tabular}{ccccc}
		\toprule
		\multirow{2}{*}{Ablation Setting} & RMSE & \multicolumn{3}{c}{AUC/(\%)}         \\ 
		\cline{3-5}
		&/px & @3px & @5px & @10px\\
		\midrule
		w/o Reconstruction & 3.31 & 16.42 & 37.32 & 65.11 \\ 
		with Reconstruction & \multirow{2}{*}{\textbf{3.17}} & \multirow{2}{*}{\textbf{17.97}} &\multirow{2}{*}{\textbf{40.05}} &  \multirow{2}{*}{\textbf{67.83}}\\
		(the complete RRSI) \\ 
		\bottomrule
	\end{tabular}
	\label{tab:ablation3}
\end{table}

\begin{table}
	\caption{Generalization Results for the RRSI Method on the MIMD dataset}
	\centering
	\begin{tabular}{ccccc}
		\toprule
		\multirow{2}{*}{Domain} & RMSE & \multicolumn{3}{c}{AUC/(\%)}         \\ 
		\cline{3-5}
		&/px & @3px & @5px & @10px\\
		\midrule
		Remote Sensing   & 1.27 & 37.78 & 54.00 & 76.33 \\
		Medical                & 2.67 & 17.78 & 39.33 & 68.33 \\ 
		Computer Vision  & 1.42 & 30.00 & 49.33 & 74.33 \\ 
		\bottomrule
	\end{tabular}
	\label{tab:generalization}
\end{table}

\textbf{Effectiveness of the joint encoding process of geometric and radiometric relations.} This ablation experiment is conducted on the OSdataset. For the feature encoding, interaction, and fusion processes involved in the RRSI feature description network, the following ablation settings are adopted: (1) Baseline: Only uses CNN to extract features and discards "Attention with Intra-Modality" and "Attention with Inter-Modality". Taking modality A in Fig. \ref{fig:framework} as an example, the features \(F'_{A,C} / F'_{A,LP}\) are used to replace \(F''_{A,C} / F''_{A,LP}\). Subsequently, the feature \(F'''_{A}\) (obtained by fusing \(F'_{A,C} / F'_{A,LP}\) via MLP) is used as the descriptor, and the same operation is applied to modality B. (2) Baseline + Attention with Intra-Modality: Based on the Baseline, "Attention with Intra-Modality" is added, and \(F'''_{A}\) and \(F'''_{B}\) are used as descriptors. (3) Baseline + Attention with Inter-Modality: Based on the Baseline, "Attention with Inter-Modality" is added. First, the CNN features \(F'_{A,C} / F'_{A,LP}\) are fused via MLP to generate feature \(F'''_{A}\), and the same MLP fusion is applied to the CNN features \(F'_{B,C} / F'_{B,LP}\) to generate feature \(F'''_{B}\). Subsequently, \(F'''_{A}\) and \(F'''_{B}\) are input into "Attention with Inter-Modality" for feature interaction, and the outputs \(F_{A}\) and \(F_{B}\) are used as descriptors. (4) Baseline + Attention with Intra-Modality \& Inter-Modality: This refers to the complete RRSI method.

The experimental results, as shown in Table \ref{tab:ablation2}, validate the effectiveness of the joint encoding of geometric and radiometric relations in the RRSI feature description network. Specifically, this network overcomes the limitation of only extracting convolutional features from local regions, enabling the process of feature encoding, interaction, and fusion for intra-modal local regions, dual-headed sampling regions under the DHRS module, and cross-modality image regions. By jointly modeling geometric and radiometric relations within a unified deep feature space, the RRSI descriptor can robustly represent common features invariant to geometric distortions and radiometric differences, meeting the core requirements of multimodal image matching.

\textbf{Effectiveness of the generative reconstruction constraint.} To explicitly evaluate the quantitative contribution of the training-phase cross-modal generative reconstruction constraint, a targeted comparative experiment is conducted on the OSdataset. We analyze two distinct configurations: (1) \textit{w/o Reconstruction}: the network is optimized conventionally using only the triplet margin loss and the similarity regularization loss, completely discarding the twin auxiliary decoders; (2) \textit{with Reconstruction}: the complete training workflow including both pixel-level spatial layout and feature-level perceptual consistency supervisions is activated.

The quantitative ablation results are summarized in Table \ref{tab:ablation3}. It can be observed that excluding the generative reconstruction constraint leads to a noticeable performance degradation, where the registration error degrades to an RMSE of 3.31 pixels, and the overall AUC metric at 3-pixel accuracy falls to 16.42\%. Symmetrically, when the bidirectional generative constraint is integrated, the highly abstracted deep features are forced to inversely resolve dense structural layouts, yielding a tighter and more discriminative modality-invariant representation. The reconstruction constraint consistently improves all metrics, reducing the RMSE to 3.17 pixels and increasing AUC@3px, AUC@5px, and AUC@10px to 17.97\%, 40.05\%, and 67.83\%, respectively. These values are consistent with the complete RRSI results reported on OSdataset in the main comparison. Because the auxiliary decoders are removed after training, these gains introduce no additional parameters or architectural changes during image matching inference, supporting the effectiveness of the decoupled multi-objective design.

\subsection{Generalization Analysis}
\label{sec:generalization}

\begin{figure*}
	\centering  %
	\subfloat[Computer Vision Image\label{fig:generalization_cv}]{\includegraphics[width=1\linewidth]{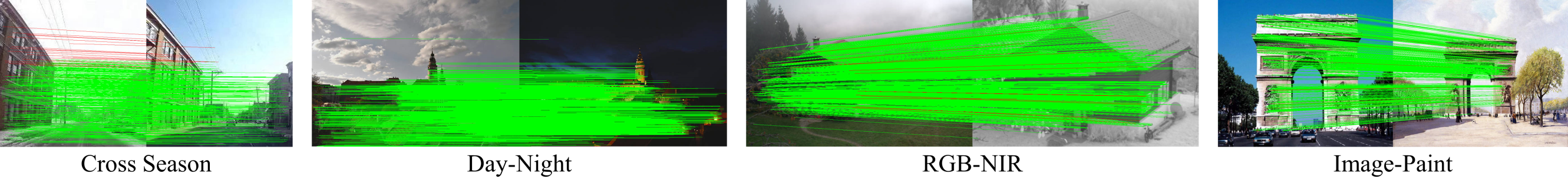}}
	
	\vspace{8pt} 
	\subfloat[Remote Sensing Image \label{fig:generalization_rs}]{\includegraphics[width=1\linewidth]{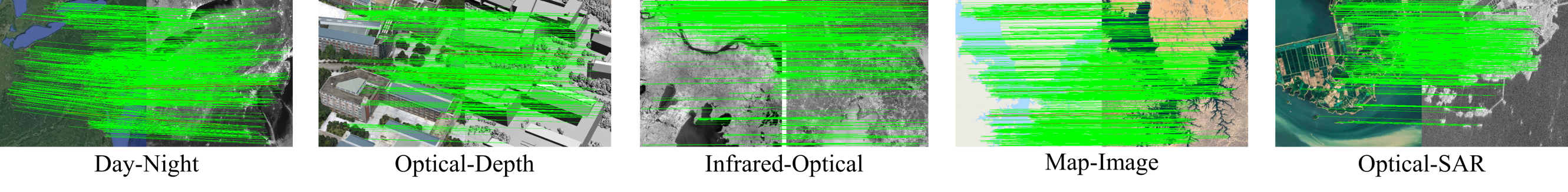}}
	
	\vspace{8pt} 
	\subfloat[Medical Image \label{fig:generalization_md}]{\includegraphics[width=1\linewidth]{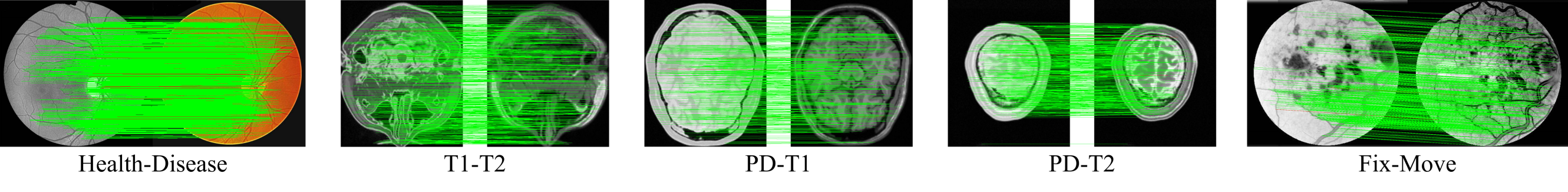}}
	
	\caption{Visual matching of the RRSI on the MIMD dataset}
	\label{fig:generalization}
	\vspace{-15pt}  %
\end{figure*}

It is widely recognized that generalization performance significantly restricts the broad adoption of deep learning methods. In this section, we conduct generalization experiments for the RRSI method. Without additional training, we use the model pre-trained on the OSdataset (optical-SAR modality), since optical-SAR is a relatively challenging type for multimodal matching with significant nonlinear radiometric differences, to evaluate the matching performance of RRSI on the MIMD dataset. We select 15 pairs of multimodal images for testing in each of the three categories, namely computer vision, remote sensing, and medical imaging, totaling 45 pairs. As presented in Table \ref{tab:generalization}, the results demonstrate that the model trained on optical-SAR data transfers reliably to unseen multimodal image types without additional training. Performance varies across domains according to the modality gap and texture clarity, while remaining stable across the remote sensing, medical imaging, and computer vision categories. Additionally, Fig. \ref{fig:generalization} presents the visualized matching results of RRSI on the MIMD dataset. The RRSI produces numerous dense and correct correspondences, which further confirms its strong generalization ability.

\section{Conclusions}
\label{sec:conclusions}
We propose the RRSI feature descriptor for multimodal image matching. Specifically, we design the DHRS module for simultaneous Cartesian and Log-Polar sampling on keypoint neighborhoods. This parallel encoding strategy lets the RRSI descriptor retain spatial structural properties while notably enhancing robustness to rotation and scale variations. Furthermore, by jointly encoding geometric and radiometric relations in a unified deep feature space, we mitigate insufficient receptive fields in keypoint neighborhoods, and achieve feature encoding, interaction, and fusion across intra-modal, dual-head sampled, and inter-modal regions. Additionally, by introducing a cross-modal generative reconstruction constraint during the training phase, the implicit features are explicitly forced to anchor modality-invariant geometric topologies without increasing inference overhead. Ultimately, the RRSI descriptor has robust invariance to geometric distortions and radiometric differences, effectively improving multimodal image matching accuracy.

Experimental results confirm that the RRSI method achieves superior or highly competitive performance compared to state-of-the-art approaches on challenging multimodal datasets (e.g., optical-infrared and optical-SAR). It supports multimodal matching over the full $[0^\circ,360^\circ]$ rotation range and at scale factors of up to four. Additionally, the RRSI’s generalization ability has been validated on additional multimodal images, including those in computer vision, remote sensing, and medical imaging, highlighting its broad practical applicability.

The current framework models the geometric relationship between images using a homography. Future work will extend RRSI to scenarios with more complex local geometric distortions and integrate its representation principles into learnable keypoint detectors and detector-free matching models, thereby further broadening its applicability.

\section*{Acknowledgments}
This work is supported by the National Key Research and Development Program of China (No. 2024YFC3015404), the National Natural Science Foundation of China (No. 42271446), and the Central Guidance for Local Science and Technology Development Fund Projects of Sichuan Province (No. 2024ZYD0084).

\bibliographystyle{IEEEtran}
\bibliography{IEEEabrv,references}
\end{document}